\documentclass[11pt]{article}

\usepackage[final]{acl}

\usepackage{times}
\usepackage{latexsym}

\usepackage[T1]{fontenc}
\usepackage[utf8]{inputenc}

\usepackage{microtype}

\usepackage{inconsolata}

\usepackage{graphicx}

\usepackage{booktabs}
\usepackage{multirow}
\usepackage{xcolor}
\usepackage{amsmath}
\usepackage{amssymb}
\usepackage{subcaption}
\usepackage{afterpage}
\usepackage{pifont}

\newcommand{\repourl}{https://github.com/kookhh0827/copy-inflation-search-agents}
\newcommand{\dataurl}{https://huggingface.co/datasets/kookhh0827/copy-inflation-search-agents}

\graphicspath{{figures/}}

\title{Beyond Confidence: Test-Time Scaling for Multi-Turn Search Agents via Retrieval Grounding}

\author{
  \textbf{Hyunho Kook\textsuperscript{1}} \quad
  \textbf{Junhyuk So\textsuperscript{2}} \quad
  \textbf{Tianyu Fu\textsuperscript{3}} \quad
  \textbf{Haizhong Zheng\textsuperscript{3}} \quad
  \textbf{Beidi Chen\textsuperscript{3}}
  \\[3pt]
  \textsuperscript{1}University of Southern California \quad
  \textsuperscript{2}Pohang University of Science and Technology (POSTECH) \\
  \textsuperscript{3}Carnegie Mellon University
  \\[3pt]
  \small{\texttt{hyunho.kook@usc.edu} \quad
         \texttt{junhyukso@postech.ac.kr} \quad
         \texttt{fuvty@outlook.com}}
  \\[1pt]
  \small{\texttt{\{haizhonz, beidic\}@andrew.cmu.edu}}
}
\begin{document}
\maketitle

% Force Figure 1 (method) to top of page 2.
\begin{figure*}[!t]
  \centering
  \includegraphics[width=0.98\textwidth]{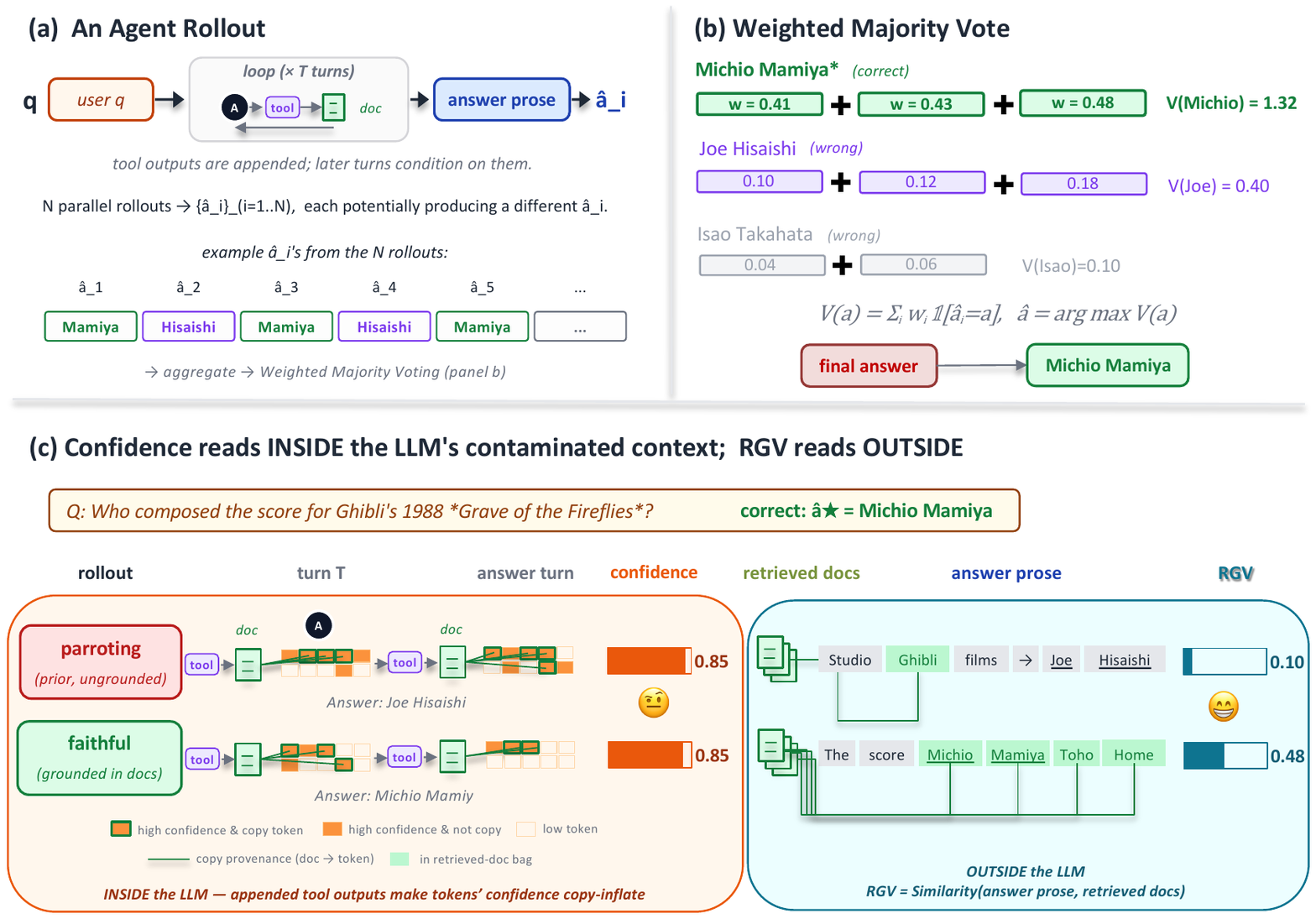}
  \caption{\textbf{Overview.}
    (a)~Each rollout is one trajectory of the ReAct-style search agent, looping over tool calls (search, page visit) and emitting a predicted answer; $N$ parallel rollouts may differ. (b)~The weighted vote sums per-rollout weights over predicted answers; argmax wins. (c)~Confidence-based voting weights each rollout by its
    token logprobs, inflated when the model parrots context. RGV (ours) weights each rollout by the lexical overlap between its answer prose and the retrieved documents---a signal read outside the contaminated context.}
  \label{fig:method}
\end{figure*}

\begin{abstract}
Confidence-based voting aggregates parallel LLM rollouts by weighting each with internal signals such as token log probabilities, and has been actively studied for single-turn reasoning. However, modern LLMs increasingly act as multi-turn search agents that retrieve and condition on external documents. In this paper, we show that confidence-based voting transfers poorly to this multi-turn setting, and identify the underlying failure reason as copy inflation: when retrieved documents are appended to an agent's context, tokens copied from those documents receive systematically inflated log probabilities. This flattens confidence scores within each question and weakens the resulting weighted vote. To address this issue, we propose Retrieval-Grounded Voting (RGV), which scores each rollout by the lexical overlap between its final answer and the documents it retrieved. By computing the signal outside the contaminated context, RGV sidesteps both token log probabilities and additional LLM calls. Across four search-agent benchmarks and five LLMs, RGV consistently outperforms confidence-based voting, with gains of up to +5.4\% accuracy and +35\% on minority-correct questions, where the correct answer appears in only 1-2 of 8 rollouts.
\end{abstract}
\section{Introduction}

Confidence-based voting, weighting parallel rollouts by an aggregate of their token-level logprobs, has been widely adopted for test-time scaling in single-turn LLM reasoning \citep{fu2025deepconf,taubenfeld2025cisc}, and can outperform majority voting on non-retrieval interactive agent tasks \citep{wang2024softsc}. But as LLMs are deployed as multi-turn search agents that condition on retrieved documents \citep{yao2022react,nakano2021webgpt,chen2025browsecompplus}, this logprob-based signal transfers poorly. Why confidence voting fails in this setting, and how to address it at the voting layer, remains largely understudied.

Existing work addresses the problem from three perspectives, but each leaves it unresolved. \emph{Confidence-based voters} \citep{fu2025deepconf,taubenfeld2025cisc,wang2024softsc} are designed for single-turn LLM settings and offer no diagnosis for search agents. \emph{Trajectory-aware aggregators} \citep{lee2026aggagent,li2025parallelmuse} sidestep the confidence signal by spending an extra LLM call per question (an aggregator on top of the parallel rollouts), adding inference cost. \emph{RL-based recalibration} \citep{xuan2026dichotomy} adjusts confidence via RL fine-tuning but leaves the source of miscalibration in search agents largely unexplained.

An ideal treatment should first \emph{diagnose} why confidence
signals degrade in search agents, then leverage that diagnosis to
design an efficient, broadly applicable voting-layer solution.

To address this gap, we make two contributions. \emph{First}, we identify and
quantify a mechanism we call \emph{copy-inflation}: once retrieved
documents are appended to the agent's context, copied tokens
receive inflated logprobs, compressing confidence scores within a
question so that weighted majority voting collapses to simple majority (\S\ref{sec:motivation}). \emph{Second}, the diagnosis points to a
solution: read the voting signal from \emph{outside the contaminated
context}. We propose \textbf{Retrieval-Grounded Voting (RGV)}:
weight each rollout by the lexical overlap between
its answer prose and the documents it retrieved
(\S\ref{sec:method}). Because the retrieval log is
environment-derived, the signal is independent of the model's hidden
state and requires no logprobs, no fine-tuning, and no extra LLM
calls.

We publicly release both our code\footnote{\begingroup\raggedright\url{\repourl}\par\endgroup} and our data\footnote{\begingroup\raggedright\url{\dataurl}\par\endgroup}: the full rollout trajectories, the per-token log-probabilities, and the judge outputs.
Across four search agent benchmarks and five LLMs, RGV beats
logprob-based confidence voting (DeepConf;
\citealp{fu2025deepconf}), by up to $+5.4\%$ accuracy and $+35\%$ on minority-correct questions. RGV at four rollouts already matches DeepConf at eight, halving the rollout budget at equal accuracy. The gains are robust to the choice of overlap metric and add negligible CPU cost.

% ======================================================================

\section{Background}
\label{sec:background}
\afterpage{%
\begin{figure*}[!t]
  \centering
  \begin{subfigure}[t]{0.32\textwidth}
    \centering
    \includegraphics[width=\linewidth]{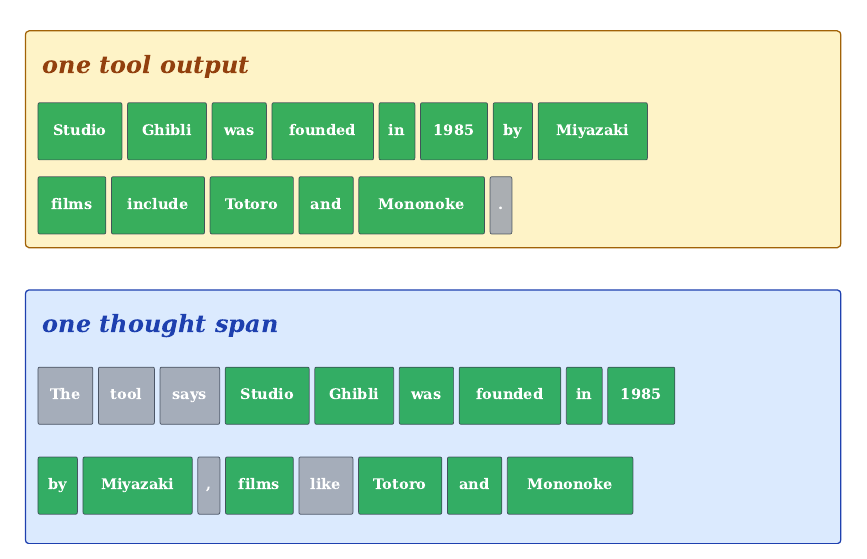}
    \caption{\textbf{What is a copy token?} A token whose surface
      form is a substring of some retrieved doc (green).}
    \label{fig:copy-primer}
  \end{subfigure}\hfill
  \begin{subfigure}[t]{0.32\textwidth}
    \centering
    \includegraphics[width=\linewidth]{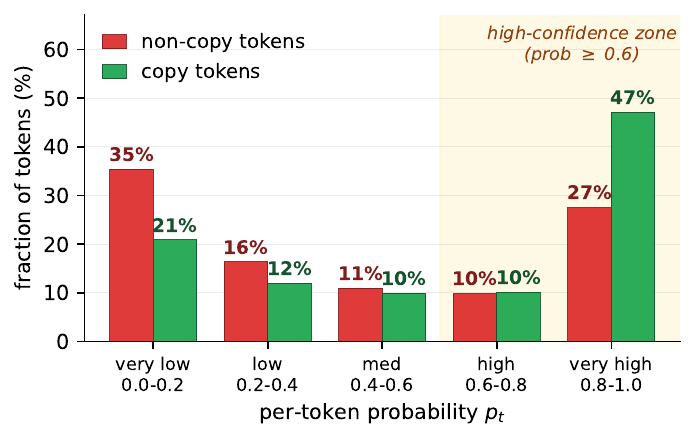}
    \caption{\textbf{Copy tokens carry inflated confidence.}
      Per-token probability: copy vs.\ non-copy.}
    \label{fig:copy-token}
  \end{subfigure}\hfill
  \begin{subfigure}[t]{0.32\textwidth}
    \centering
    \includegraphics[width=\linewidth]{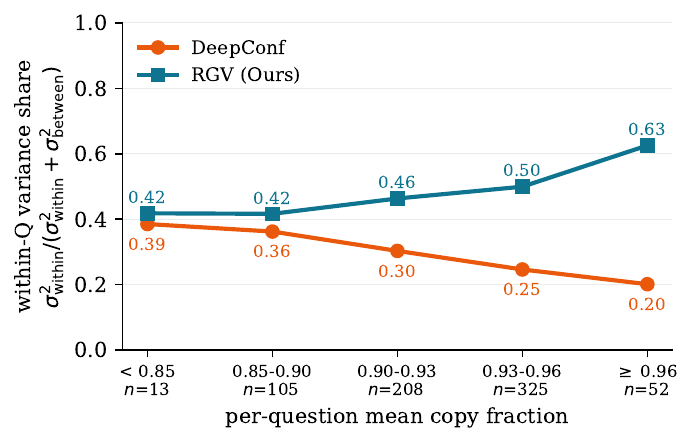}
    \caption{\textbf{Copy-heaviness flattens DeepConf within a
      question.} Within-$Q$ score-variance share vs.\ per-$Q$
      copy fraction.}
    \label{fig:copy-compress}
  \end{subfigure}
  \caption{\textbf{Copy-inflation.} (BrowseComp-Plus, Tongyi-DeepResearch)
    (\subref{fig:copy-primer}) Defines the copy-token primitive:
    the doc is already in context, and the model parrots the tool outputs.
    (\subref{fig:copy-token}) On generated tokens, copy tokens cluster at prob ${\geq}0.6$
    ($57\%$ vs.\ $37\%$ for non-copy), a $+0.50$ mean logprob gap.
    (\subref{fig:copy-compress}) The $y$-axis is the within-question
    share of total DeepConf score variance, i.e.\ how much of the
    score's spread comes from differences among rollouts of the
    \emph{same} question. As per-$Q$ mean copy fraction grows,
    DeepConf's share drops $0.39{\to}0.20$ (same-$Q$ rollouts get
    nearly the same score, so weighted voting cannot separate
    them); for comparison, RGV (our method, \S\ref{sec:method})
    rises $0.42{\to}0.63$. Median rollout copies
    $93\%$ of its content tokens; $85\%$ of questions have mean
    copy fraction ${\geq}0.90$.}
  \label{fig:copy}
\end{figure*}
\begin{figure*}[!t]
  \centering
  \begin{subfigure}[t]{0.38\textwidth}
    \centering
    \includegraphics[width=\linewidth]{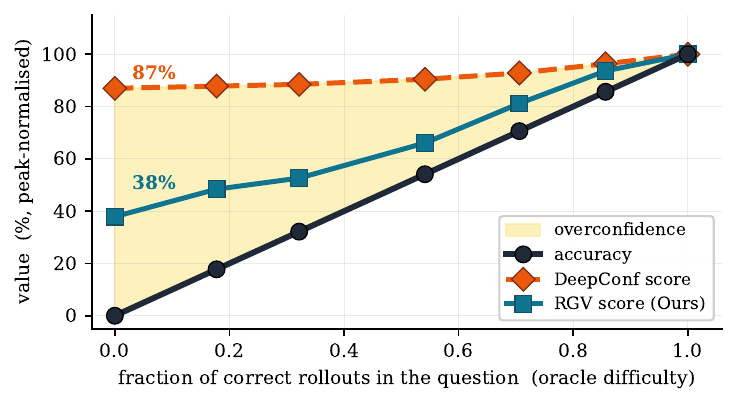}
    \caption{\textbf{DeepConf is overconfident.}
      Mean per-question DeepConf score stays at $87\%$ of peak even
      on questions where every rollout is wrong.}
    \label{fig:motivation-overconf}
  \end{subfigure}\hfill
  \begin{subfigure}[t]{0.28\textwidth}
    \centering
    \includegraphics[width=\linewidth]{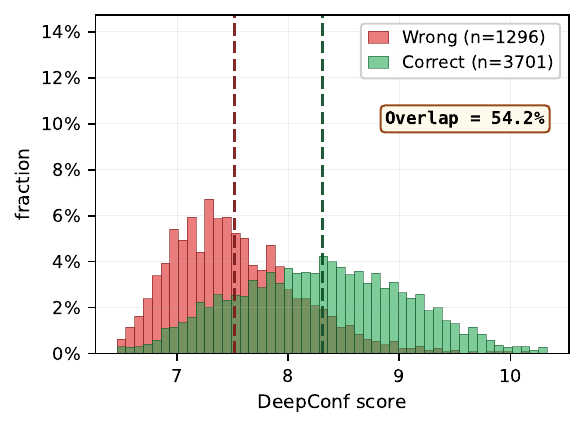}
    \caption{\textbf{DeepConf.} Per-rollout score density split by
      judge correctness; the two populations overlap heavily.}
    \label{fig:motivation-dc}
  \end{subfigure}\hfill
  \begin{subfigure}[t]{0.28\textwidth}
    \centering
    \includegraphics[width=\linewidth]{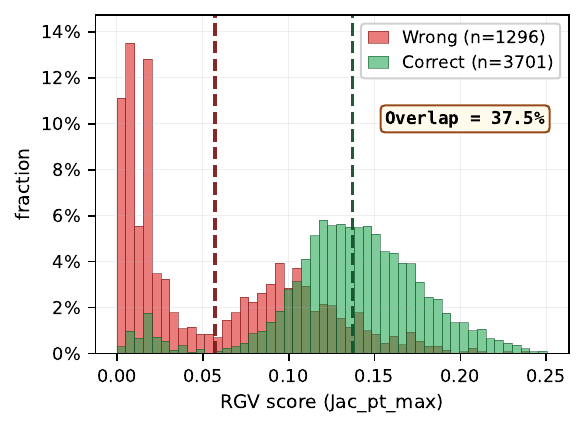}
    \caption{\textbf{RGV (Ours; \S\ref{sec:method}).} Same split as
      (b), scored by our method (overlap $37.5\%$).}
    \label{fig:motivation-tgv}
  \end{subfigure}
  \caption{\textbf{Consequence at the rollout level}
    (BrowseComp-Plus, Tongyi-DeepResearch). (\subref{fig:motivation-overconf}) Stratifying
    questions by oracle difficulty (fraction of correct rollouts in
    the question) reveals DeepConf is nearly flat: it gives
    impossible questions ($87\%$ of peak), almost the same confidence
    it gives easy ones. (\subref{fig:motivation-dc})
    DeepConf's per-rollout correct/wrong distributions overlap
    heavily ($54.2\%$). (\subref{fig:motivation-tgv}) For
    comparison: RGV (our method, \S\ref{sec:method}) on the same split.}
  \label{fig:motivation}
\end{figure*}
}

\paragraph{Multi-turn search agent.}
We study ReAct-style agents
\citep{yao2022react,schick2023toolformer}
whose tools retrieve text documents and whose final answer is
derived from the retrieved content. We call these \emph{multi-turn search
agents}, or \emph{search agents} for short. Such an agent
interleaves reasoning and tool calls under a budget
of $T$ turns. With context $c_0 = q$, at turn $k$ the policy
$\pi_\theta$ samples $(u_k, x_k) \sim \pi_\theta(\cdot \mid c_{k-1})$,
where $u_k$ is the model's thought span at turn $k$ and $x_k$ is either a tool
invocation (e.g.\ \texttt{search} to query a search engine, or
\texttt{visit} to fetch a web page) or the final
answer. On a tool call the environment returns an observation $o_k$
and the context grows by
$c_k = c_{k-1} \Vert u_k \Vert x_k \Vert o_k$. Retrieved snippets
are thus appended to the same context the model conditions on at
later turns.

\paragraph{Rollout.}
A \emph{rollout} $t_i$ is one trajectory of this loop, terminating
on a final answer or budget exhaustion. We sample $N$ rollouts
$\{t_1,\dots,t_N\}$ in parallel from $q$. The rollout's record
contains every tool call and its response, every thought span
the model emitted, and a final \emph{answer turn}
whose output is one last thought span, followed by the predicted
answer string. Three names from this record recur below: the
predicted answer $\hat{a}_i$, the \emph{answer prose} $P_i$ (the final answer the model produces
on the answer turn, excluding its internal chain-of-thought), and the retrieval log $\mathcal{D}_i = \{d_1,\dots,d_{m_i}\}$
(the documents the agent fetched).

\paragraph{Weighted majority vote.}
The $N$ rollouts may give different answers, so a voting rule
returns a single final answer. Each rollout is scored by a
non-negative weight $w_i$, and the chosen answer is the one whose
rollouts carry the largest total weight:
\[
  V(a) = \sum_{i=1}^{N} w_i \, \mathbb{I}(\hat{a}_i = a),
  \qquad
  \hat{a} = \arg\max_a V(a).
\]
Different choices of $w_i$ give different methods. \emph{Simple
majority} ($w_i\!\equiv\!1$) just counts rollouts per answer.
\emph{DeepConf} \citep{fu2025deepconf}, our confidence baseline,
reduces per-token confidences $C_t = -\tfrac{1}{k}\sum_{j=1}^{k}\log
p_j$ (top-$k$ averaged log-prob, negated; higher means more
peaked) over the tokens the model produces during the rollout to a single $w_i$
via a sliding-window aggregation. We use the \emph{Lowest-Group} reduction (sliding-window minimum)
at window size $W{=}1024$ as the headline; the full sweep over
reductions and window sizes is in Appendix~\ref{app:dc-variants}. Our method (\S\ref{sec:method})
replaces $w_i$ with a direct grounding check between the answer
prose $P_i$ and the retrieved docs $\mathcal{D}_i$.

% ======================================================================

\section{Motivation}
\label{sec:motivation}

When the only input is the question, DeepConf \citep{fu2025deepconf}, a logprob-based confidence voter (\S\ref{sec:background}) is well-motivated: peaked next-token posteriors correlate with correctness \citep{kadavath2022knowwhattheyknow,tian2023justaskcalibrated}. In a search agent, however, that correlation breaks down because the tokens DeepConf reads are systematically inflated by the retrieved docs appended to the context. We document this on BrowseComp-Plus rollouts from Tongyi-DeepResearch in three observations (Fig.~\ref{fig:copy}) and trace the consequence at the rollout level (Fig.~\ref{fig:motivation}).

\paragraph{Tool outputs in context inflate logprobs.}
A search agent appends every retrieved snippet and visited page to its context window; when it later writes prose, any token it copies from those appended documents has very high $p(t_k \mid \text{context})$ because the document is right there. The token's logprob says the \emph{copy step} is well-calibrated; it does not say the \emph{copied token} is the right answer. Confidence-based voting cannot tell the difference between a rollout that copied the right entity and one that copied an irrelevant one.

\emph{(i) Copy tokens carry inflated confidence}
(Fig.~\ref{fig:copy-token}). Recall (\S\ref{sec:background}) that DeepConf reads the tokens the model produces during the rollout. For each such token we check whether its surface form is a substring of any retrieved doc the agent fetched on this rollout; copy tokens have a mean logprob $+0.50$ nats higher than non-copy tokens (730k tokens from a subset of rollouts where per-token logprobs were re-collected). The signal DeepConf reads is thus contaminated by the appended tool outputs regardless of what those tools returned. The gap is robust to the choice of copy-token definition (Appendix~\ref{app:copy}).

\emph{(ii) At the rollout level, copy-inflation flattens
DeepConf scores within a question} (Fig.~\ref{fig:copy-compress}). We measure DeepConf's \emph{within-question share} of total score variance, i.e.\ how much of the score's spread comes from differences \emph{among rollouts of the same question}, rather than between questions. Weighted majority voting needs this within-question spread to prefer one rollout over another; if it vanishes, the vote degenerates into plain simple-majority. As copy fraction grows, DeepConf's within-question share shrinks $0.39\to0.20$: same-question rollouts get nearly the same DeepConf score and the weight tiebreaker disappears, precisely on the copy-heavy questions where rollouts disagree most.

\emph{(iii) And this is not an edge case; copy ratios are very high}. Across our rollouts, the median copies $93\%$ of its content tokens from the docs it retrieved, $81\%$ copy at least $90\%$, and $85\%$ of questions have a mean copy fraction ${\geq}0.90$ (Fig.~\ref{fig:copy-compress}, $x$-axis). The two failure modes above therefore apply to nearly the entire benchmark, not a corner case. While copy-inflation can in principle arise whenever tool outputs enter the agent's context; multi-turn search agents are the dominant regime. Two interventions confirm the direction of causation rather than mere association: masking the copied tokens restores DeepConf's within-question spread without restoring its discriminative power, and removing the documents from context collapses copied-token log-probabilities about twice as much as non-copy ones (Appendix~\ref{app:interventions}).

\paragraph{Consequence: the weighted vote no longer tracks
correctness.} Figure~\ref{fig:motivation} makes the rollout-level consequence concrete. \emph{(a)} Stratifying questions by oracle difficulty (the fraction of their $N$ rollouts that turn out correct), DeepConf's mean per-question score remains at $87\%$ of its peak even on questions where \emph{every} rollout is wrong: DeepConf cannot tell an impossible question from a solvable one, so the cluster-weighted vote inherits this miscalibration at the voting layer; The pattern replicates across every benchmark and model we test (Appendix~\ref{app:cross-dataset}). \emph{(b)} The same pattern holds at the rollout level: DeepConf's per-rollout score distributions for correct vs.\ wrong rollouts overlap heavily (overlap ${=}54.2\%$), as copy-inflation lifts wrong rollouts into the same high-confidence zone as correct ones, leaving the confidence weight nothing to discriminate.
% Panel (c) shows the same split scored by our method (RGV, \S\ref{sec:method}); we defer its interpretation to that section.

\paragraph{The principle.}
The shared cause of failure modes is that the voting signal is computed \emph{inside} the contaminated context, not outside it. Robust voting therefore requires a signal whose measurement target is independent of the model's hidden state; something derived from outside the model. The next section turns this principle into a method.

% ======================================================================

\section{From principle to method: Retrieval-Grounded Voting}
\label{sec:method}

To address the copy-inflation problem identified in
\S\ref{sec:motivation}, we propose reading the voting signal from
\emph{outside the contaminated context}, specifically, from the
retrieval log. The key intuition, shared with faithfulness evaluation
in summarisation and RAG
\citep{maynez2020faithfulness,min2023factscore,es2024ragas}, is
that a rollout is more trustworthy when its final answer is
lexically anchored in the documents it retrieved.

\paragraph{The RGV score.}
For each rollout $i$ on a given question, we have its answer prose
$P_i$ (defined in \S\ref{sec:background}: the answer turn's output
text) and the \emph{set}
$\mathcal{D}_i = \{d_1, \dots, d_{m_i}\}$ of documents the agent
retrieved during that rollout (one $d_j$ per \texttt{search} or
\texttt{visit} call). We map a text span $X$ to a token set
$\mathcal{T}(X)$ using a fixed tokenisation/normalisation rule
(Appendix~\ref{app:impl}). The rollout's RGV weight is the
maximum \emph{prose-recall}, the fraction of the answer-prose
tokens that are anchored in some retrieved document,
\[
  w_i^{\text{RGV}}
  \;=\;
  \max_{d \in \mathcal{D}_i}
  \frac{|\mathcal{T}(P_i) \cap \mathcal{T}(d)|}
       {|\mathcal{T}(P_i)|}.
\]

This is the standard grounding primitive of the faithfulness
literature (\S\ref{sec:related}), instantiated with two design
choices, max-over-docs and answer-side normalisation, which we
motivate next. Appendix~\ref{app:lex-metrics} shows that these choices
matter only at the margin, while the signal \emph{source} carries the
bulk of the gain.

\paragraph{Why max-over-docs.}
We score against the single best-matching document rather than
the union of all retrieved documents to avoid dilution: as the
retrieval bag grows, an irrelevant-doc union washes out a strong
match to the truly supporting passage. This ``take an extreme,
not an average'' design mirrors local-extreme confidence
reductions in DeepConf \citep{fu2025deepconf}, which prefer the
most-uncertain window-group over an average across all windows.

\paragraph{Why the prose-side denominator.}
With the numerator fixed at
$|\mathcal{T}(P_i)\cap\mathcal{T}(d)|$, the design choice is
the denominator. Symmetric Jaccard \citep{jaccard1912index} uses
$|\mathcal{T}(P_i)\cup\mathcal{T}(d)|$, which over-penalises
when the retrieved doc is long, exactly the regime we target.
Normalising by $|\mathcal{T}(P_i)|$ keeps the score
length-invariant on the doc side: a well-anchored answer is not
punished for retrieving a long supporting passage. Other
overlap functions \citep{lin2004rouge,robertson2009bm25}
yield similar headline numbers (Appendix~\ref{app:lex-metrics});
the gain comes from the signal source, not the choice of
denominator.

\paragraph{Voting rule.}
We plug $w_i \!=\! w_i^{\text{RGV}}$ into the weighted majority
vote of \S\ref{sec:background}; answer strings are clustered after
a light normalisation (lowercase, whitespace collapse,
leading-article strip) and the cluster with the largest summed
weight wins.

\paragraph{What is and is not computed inside the model's context.}
The token set $\mathcal{T}(P_i)$ is, of course, generated by the
model. The token set $\mathcal{T}(d)$ is read directly off the
retrieval log: it is what the environment returned, regardless of
how the model used it. Even though the docs were appended to the
model's context, their token set as a \emph{measurement target}
does not depend on the model's hidden state, and a rollout cannot
inflate its RGV weight through verbosity or self-assuredness; RGV
asks only whether the answer prose is anchored in the documents
the rollout actually retrieved.

% ======================================================================

\section{Experimental setup}
\label{sec:setup}

\paragraph{Benchmarks.}
We evaluate on four multi-turn search agent benchmarks where the
agent gathers evidence from retrieved documents and derives a
grounded answer: BrowseComp-Plus
\citep{chen2025browsecompplus} (fixed-corpus retrieval),
BrowseComp \citep{wei2025browsecomp} (open-web browsing), GAIA
\citep{mialon2023gaia} (general-assistant tasks dominated by web
search), and FRAMES \citep{krishna2024frames} (multi-hop retrieval
requiring synthesis across multiple Wikipedia articles). For each
benchmark, we randomly sample up to $150$ questions, held constant
across models. Licences and intended-use terms for every benchmark, corpus and model are listed in Appendix~\ref{app:licences}.

\paragraph{Models.}
We evaluate five LLMs across the four benchmarks
above: gpt-oss-120b \citep{openai2025gptoss}, MiniMax-M2.7
\citep{minimax2025m1}, GLM-5.1 \citep{zai2026glm5},
Kimi-K2.5 \citep{moonshot2025kimik2}, and Tongyi-DeepResearch
\citep{tongyideepresearch2025}. The resulting benchmark $\times$
model coverage is reported in Table~\ref{tab:headline}.

\paragraph{Rollout protocol.}
The agent is given two tools per benchmark:
\texttt{search} (issue a text query, receive ranked snippets) and
\texttt{get\_doc} (fetch a full document by ID) on BrowseComp-Plus
(fixed-corpus retrieval), and \texttt{search} and \texttt{visit}
(fetch and render a URL) on the other three benchmarks. We sample
$N{=}8$ rollouts per question. Full
hyperparameters, tool truncation rules, and the force-final /
recovery procedure are in Appendix~\ref{app:impl}.

\paragraph{Judge.}
A single judge is used across all datasets:
Qwen3-32B at temperature $0$, applied with the official
BrowseComp-Plus grading template
\citep{chen2025browsecompplus}. The judge sees only the
question, the gold answer, and the rollout's final answer; it
does not see voting outcomes or other rollouts.

\paragraph{Voting protocol.}
All voting methods plug a per-rollout weight $w_i$ into the
weighted majority vote of \S\ref{sec:background}; answer
strings are clustered under the strict normalisation of
\S\ref{sec:method}. Methods compared:
\emph{Simple-majority} (\textsc{SM}, $w_i\!\equiv\!1$);
\emph{DeepConf} (\textsc{DC}; Lowest-Group reduction with window
$W{=}1024$, chosen on the
full reduction $\times$ window-size grid in
Appendix~\ref{app:dc-variants});
\emph{RGV} (\S\ref{sec:method}).
Two reference columns in Table~\ref{tab:headline} are non-voting:
\emph{single} = mean per-rollout accuracy, and
\emph{oracle} = any-rollout-correct upper bound.

% ======================================================================

\section{Results}
\label{sec:results}

\begin{table*}[t]
  \centering
  \small
  \setlength{\tabcolsep}{4.0pt}
  \renewcommand{\arraystretch}{1.05}
  \begin{tabular}{l l ccccc}
    \toprule
    \textbf{Dataset} & \textbf{Model} &
      Single Avg. & Simple Maj. & DeepConf &
      \textbf{RGV (Ours)} & Oracle \\
    \midrule
    \multirow{5}{*}{BrowseComp-Plus}
      & Tongyi-DeepResearch & 51.6 & 62.0   & 65.7   & \textbf{71.1} & 74.2 \\
      & OSS-120B            & 47.3 & 53.3   & 54.7   & \textbf{56.0} & 66.7 \\
      & MiniMax-M2.7        & 67.8 & 75.3   & 75.3   & \textbf{78.7} & 83.3 \\
      & Kimi-K2.5           & 80.0 & 80.7   & 80.7   & \textbf{81.3} & 88.7 \\
      & GLM-5.1             & 74.9 & 80.7 & 80.7 & \textbf{82.0} & 90.0 \\
    \cmidrule(lr){1-7}
    \multirow{5}{*}{GAIA}
      & Tongyi-DeepResearch & 70.0 & 79.6 & 79.6 & \textbf{80.6}   & 92.2 \\
      & OSS-120B            & 60.4 & 69.9 & 68.9 & \textbf{71.8}   & 83.5 \\
      & MiniMax-M2.7        & 80.3 & 87.4 & 88.3 & \textbf{93.2}   & 97.1 \\
      & Kimi-K2.5           & 73.1 & 78.6 & 79.6 & \textbf{81.6}   & 92.2 \\
      & GLM-5.1             & 83.3 & 87.4 & 87.4 & \textbf{89.3}   & 92.2 \\
    \cmidrule(lr){1-7}
    \multirow{5}{*}{BrowseComp}
      & Tongyi-DeepResearch & 41.2 & 52.0 & 52.7 & \textbf{54.0}   & 68.7 \\
      & OSS-120B            & 22.7 & 30.0 & 29.3 & \textbf{34.0}   & 41.3 \\
      & MiniMax-M2.7        & 37.1 & 40.7 & 46.0 & \textbf{50.7}   & 59.3 \\
      & Kimi-K2.5           & 35.0 & 52.8 & 56.6 & \textbf{58.5}   & 73.6 \\
      & GLM-5.1             & 57.8 & 72.6 & 73.7 & \textbf{78.9}   & 92.6 \\
    \cmidrule(lr){1-7}
    \multirow{5}{*}{FRAMES}
      & Tongyi-DeepResearch & 84.0 & 87.3 & 88.0 & \textbf{88.0}   & 94.7 \\
      & OSS-120B            & 78.7 & 82.0 & 82.7 & \textbf{85.3}   & 91.3 \\
      & MiniMax-M2.7        & 86.5 & 88.7 & 89.3 & \textbf{89.3}   & 94.0 \\
      & Kimi-K2.5           & 85.7 & 88.0 & 88.7 & \textbf{89.3}   & 97.3 \\
      & GLM-5.1             & 88.3 & 90.0 & 90.0 & \textbf{90.7}   & 98.0 \\
    \bottomrule
  \end{tabular}
  \caption{\textbf{Headline voting accuracies (\%)} across 4
    multi-turn search agent benchmarks $\times$ 5 LLMs,
    $N{=}8$ rollouts per question.}
  \label{tab:headline}
\end{table*}

% Table~\ref{tab:headline} summarises the results. RGV is the
% top-scoring rule on every filled cell; gains are largest where
% single-rollout accuracy is lowest (e.g.\ $+4.7\%$ on BrowseComp
% OSS-120B, $+5.4\%$ on BrowseComp-Plus Tongyi) and remain positive
% even where DeepConf already saturates (e.g.\ $+0.6\%$ on
% BrowseComp-Plus Kimi-K2.5). The gains are robust to the within-cell
% 3-fold split (Appendix~\ref{app:variance}).

% For the mechanism and ablation analyses that follow
% (\S\ref{sec:exp-why}--\S\ref{sec:exp-scaling}), we anchor in
% Tongyi-DeepResearch on BrowseComp-Plus ($N{=}8$, 830 questions, judged by Qwen3-32B). BrowseComp-Plus retrieves
% from a fixed corpus rather than the open web, removing search-engine
% noise; Tongyi-DeepResearch is a dedicated search agent,
% representative of the model class this paper targets.

Table~\ref{tab:headline} shows that RGV is best in all 20 benchmark$\times$model cells. The pattern is consistent with the copy-inflation story of \S\ref{sec:motivation}: whenever confidence weights lose within-question spread, grounding is the remaining discriminator.

Gains are largest where voting is hardest: low single-rollout accuracy (e.g.\ $+5.4\%$ on BrowseComp-Plus Tongyi at $51.6\%$ single) and noisy open-web retrieval (BrowseComp, up to $+6.2\%$). They shrink where majority already saturates (FRAMES; $\geq 84\%$ single), but remain nonzero even when DeepConf matches simple majority (e.g.\ GLM-5.1 on FRAMES: $90.0\%$ vs.\ $90.7\%$), suggesting that lexical grounding adds information beyond both count (simple-majority votes) and logprob confidence. All margins are stable under within-cell 3-fold splits (Appendix~\ref{app:variance}).

We probe the boundaries of this result in three appendices: degraded retrieval, prompt and answer-format sensitivity, and selection versus voting, in Appendices~\ref{app:weak-retrieval}, \ref{app:prompt-format} and~\ref{app:bon}.

For the mechanism and ablation analyses that follow (\S\ref{sec:exp-why}\,--\,\S\ref{sec:exp-scaling}), we anchor on Tongyi-DeepResearch on BrowseComp-Plus ($N{=}8$, 830 questions), where retrieval comes from a fixed corpus (no search-engine noise) and the agent is representative of the class we target.

% ======================================================================
\subsection{Is the signal genuine grounding?}
\label{sec:exp-why}

% \S\ref{sec:method} claims that prose-vs-retrieval overlap captures
% grounding quality, but the score could instead track length, word
% frequency, or some other confound. At the rollout level, RGV's overlap score predicts judge-marked correctness
% with ROC AUC \citep{fawcett2006roc} $0.908$, vs.\ $0.837$ for
% DeepConf ($+0.071$). We validate this advantage from three angles:
% (i) what tokens carry the signal, (ii) whether it survives controlled
% perturbations, and (iii) whether it correlates with an independent
% measure of retrieval quality.

\S\ref{sec:method} interprets overlap as grounding, but a spurious
alternative is that it rewards verbosity or common tokens. If so,
RGV would succeed by accident, and overlap could not support the
copy-inflation account of \S\ref{sec:motivation}. We therefore test
whether the score reflects genuine retrieval grounding.

At the rollout level, RGV separates judge-correct from judge-wrong
trajectories better than DeepConf (ROC AUC \citep{fawcett2006roc}
$0.908$ vs.\ $0.837$; $+0.071$). To explain \emph{why}, we analyse
(i) which token classes carry the gap, (ii) controlled shuffles and
nulls, and (iii) correlation with an independent retrieval-quality
measure.

\begin{figure}[t]
  \centering
  \begin{subfigure}[t]{0.485\columnwidth}
    \centering
    \includegraphics[width=\linewidth]{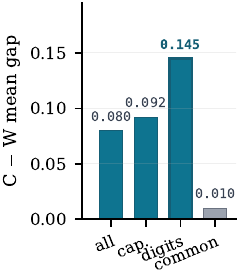}
    \subcaption{rare-entity matching}
    \label{fig:why-a}
  \end{subfigure}
  \hfill
  \begin{subfigure}[t]{0.485\columnwidth}
    \centering
    \includegraphics[width=\linewidth]{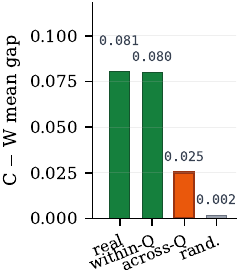}
    \subcaption{null controls}
    \label{fig:why-b}
  \end{subfigure}
  \caption{\textbf{Why RGV works} (BrowseComp-Plus Tongyi-DeepResearch).
    $y$-axis: C--W mean gap (mean RGV score of correct rollouts
    minus that of wrong rollouts).
    (\subref{fig:why-a}) Rare-entity tokens carry the signal.
    (\subref{fig:why-b}) The signal survives a within-Q doc shuffle,
    collapses $3\times$ across questions, and $53\times$ under a
    random-token null.}
  \label{fig:why}
\end{figure}

% \paragraph{Rare-entity matching drives the signal.}
% Figure~\ref{fig:why}(\subref{fig:why-a}) decomposes the
% \emph{C--W mean gap}, the difference in mean RGV score between
% judge-correct~(C) and judge-wrong~(W) rollouts, by token class.
% Digit tokens (years, codes,
% identifiers) give the largest gap ($0.145$), capitalised entity
% names follow ($0.092$), and common content words contribute little
% ($0.010$). RGV's voting power comes from rare-entity matching
% between the answer prose and the retrieved evidence, consistent
% with the rare-token weighting in classical IR
% \citep{robertson2009bm25} and atomic-faithfulness metrics
% \citep{min2023factscore}.

\paragraph{Rare-entity matching drives the signal.}
Figure~\ref{fig:why}(\subref{fig:why-a}) decomposes the
\emph{C--W mean gap}, the difference in mean RGV score between
judge-correct~(C) and judge-wrong~(W) rollouts, by token class.
Digit tokens (years, codes,
identifiers) give the largest gap ($0.145$), capitalised entity
names follow ($0.092$), and common content words contribute little
($0.010$). The pattern is intuitive: common words appear in every
rollout's prose regardless of correctness, so they carry no
discriminative power; rare entities appear only when the rollout
actually retrieved the relevant document. This aligns with the
rare-token weighting in classical IR \citep{robertson2009bm25}
and atomic-faithfulness metrics \citep{min2023factscore}.

% \paragraph{Null controls confirm question-specific grounding.}
% To distinguish genuine grounding from incidental overlap, we
% construct three counterfactuals
% (Figure~\ref{fig:why}(\subref{fig:why-b})). The C--W mean gap is $0.080$. (i)~A \emph{within-question} doc shuffle (re-pair
% each rollout's prose with another rollout's docs from the same
% question) preserves the gap at $0.080$, since same-question
% rollouts share relevant retrievals. (ii)~An \emph{across-question}
% shuffle (re-pair with a different question's docs entirely) collapses
% the gap to $0.025$ ($3\times$ drop), confirming the signal depends on
% the question-document match. (iii)~A length-matched random-token
% null collapses it to $0.002$ ($53\times$), ruling out length as a
% confound.

\paragraph{Null controls confirm question-specific grounding.}
To distinguish genuine grounding from incidental overlap, we
construct three counterfactuals
(Figure~\ref{fig:why}(\subref{fig:why-b})). The C--W mean gap is $0.080$. (i)~A \emph{within-question} doc shuffle (re-pair
each rollout's prose with another rollout's docs from the same
question) preserves the gap at $0.080$, since same-question
rollouts share relevant retrievals. (ii)~An \emph{across-question}
shuffle collapses
the gap to $0.025$ ($3\times$ drop). (iii)~A length-matched
random-token null collapses it to $0.002$ ($53\times$). Together,
they rule out the two most plausible confounds: (ii) shows
the signal is not generic text similarity but depends on the
question-document match; (iii) shows it is not an artifact of answer
length.

% \paragraph{Gold-document validation.}
% BrowseComp-Plus provides gold supporting documents per question
% (mean $|\mathcal{G}|{=}2.9$). For each rollout we compute gold
% recall, $|\mathcal{D}_i \cap \mathcal{G}|/|\mathcal{G}|$.
% Judge-correct rollouts have mean gold recall $0.94$, vs.\ $0.31$
% for judge-wrong (Pearson $r{=}0.72$). RGV scores correlate with
% gold recall at $r{=}0.57$, vs.\ $0.36$ for DeepConf; the lexical
% signal tracks genuine retrieval relevance, not merely prose
% verbosity. \S\ref{sec:exp-stratification} shows this gap
% concentrates on the minority-correct questions where the voting
% advantage is largest.

\paragraph{Gold-document validation.}
BrowseComp-Plus provides gold supporting documents per question
(mean $|\mathcal{G}|{=}2.9$). For each rollout we compute gold
recall, $|\mathcal{D}_i \cap \mathcal{G}|/|\mathcal{G}|$.
Judge-correct rollouts have mean gold recall $0.94$, vs.\ $0.31$
for judge-wrong (Pearson $r{=}0.72$). RGV scores correlate with
gold recall at $r{=}0.57$, vs.\ $0.36$ for DeepConf. In other
words, a high RGV score is a reliable indicator that the rollout
found the right documents, not merely that it produced verbose
prose. This closes the validation loop: RGV works because it
measures retrieval quality, the very property that copy-inflation
prevents DeepConf from reading.

% ======================================================================
\subsection{Where the advantage concentrates}
\label{sec:exp-stratification}

\begin{figure}[t]
  \centering
  \includegraphics[width=\columnwidth]{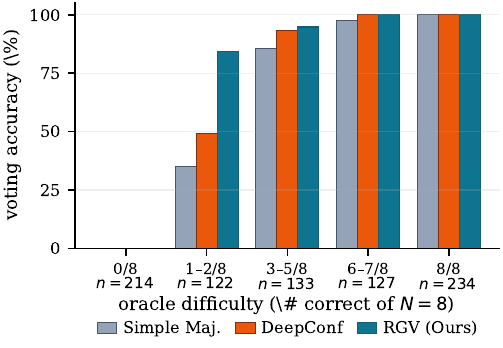}
  \caption{\textbf{Voting accuracy by question difficulty}
    (\# correct of $N{=}8$) on BrowseComp-Plus -DeepResearch. Most of the gap concentrates on the $1{-}2/8$ bucket.}
  \label{fig:stratification}
\end{figure}

% \S\ref{sec:motivation} showed that copy-inflation compresses
% DeepConf's within-question spread on exactly the questions where
% rollouts disagree. If RGV's grounding signal is immune to this
% compression, its advantage should concentrate on those same
% questions.

% Figure~\ref{fig:stratification} stratifies the 830 questions by
% oracle difficulty (the number of the eight rollouts that the judge
% marks correct). On easy questions ($\geq 3/8$ correct) all methods
% converge. Most of the $+5.4\%$ accuracy gap between RGV and DeepConf
% concentrates on the $122$ \emph{minority-correct} ($1{-}2/8$)
% questions, where one or two of eight rollouts found the right
% answer but the majority did not: RGV scores $84.4\%$ vs.\ DeepConf's
% $49.2\%$ ($+35.2\%$). DeepConf fails here because copy-inflation
% (\S\ref{sec:motivation}) gives the wrong majority nearly the same
% confidence as the correct minority; RGV breaks the tie through
% grounding. This is also the bucket where the gold-doc recall gap of \S\ref{sec:exp-why} is largest (RGV-argmax $84\%$ vs.\ DeepConf $71\%$, $+13\%$),
% confirming that the voting advantage traces back to retrieval
% quality. RGV's value is concentrated precisely where practitioners
% need it most: the ambiguous questions where rollouts disagree and
% the majority answer may be wrong.

The copy-inflation analysis of \S\ref{sec:motivation} makes a
natural prediction: because copy-inflation compresses DeepConf's
within-question spread on exactly the questions where rollouts
disagree, RGV's advantage should concentrate on those same
hard-to-vote questions. We test this prediction directly.

Figure~\ref{fig:stratification} stratifies the 830 questions by
oracle difficulty (the number of the eight rollouts that the judge
marks correct). On easy questions ($\geq 3/8$ correct) all methods
converge. Most of the accuracy gap between RGV and DeepConf
concentrates on the $122$ \emph{minority-correct} ($1{-}2/8$)
questions, where one or two of eight rollouts found the right
answer but the majority did not: RGV scores $84.4\%$ vs.\ DeepConf's
$49.2\%$ ($+35.2\%$). DeepConf fails here because copy-inflation
(\S\ref{sec:motivation}) gives the wrong majority nearly the same
confidence as the correct minority; RGV breaks the tie through
grounding. This is also the bucket where the gold-doc recall gap of \S\ref{sec:exp-why} is largest (RGV-argmax $84\%$ vs.\ DeepConf $71\%$, $+13\%$),
confirming that the voting advantage traces back to retrieval
quality. RGV's value is concentrated precisely where it is needed: the ambiguous questions where rollouts disagree and
the majority answer may be wrong.

% ======================================================================
\subsection{Scaling, cost, and design choices}
\label{sec:exp-scaling}

\begin{figure}[t]
  \centering
  \includegraphics[width=\columnwidth]{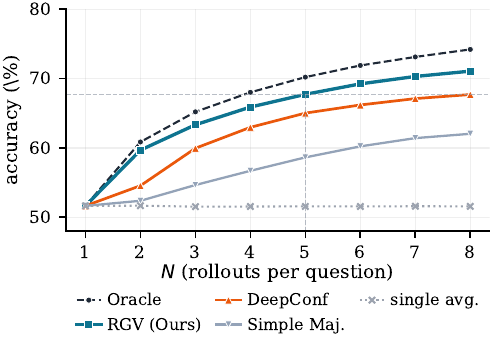}
  \caption{\textbf{Accuracy vs.\ rollout budget $N$} on
    BrowseComp-Plus with Tongyi-DeepResearch. RGV is uniformly above DeepConf.}
  \label{fig:scaling}
\end{figure}

Two practical questions remain: does RGV's advantage persist under
tighter rollout budgets, and is the max-over-docs design of
\S\ref{sec:method} the right choice?

\paragraph{Scaling with $N$.}
% Sub-sampling the eight rollouts to $N \in \{1,\dots,8\}$ at $60$
% random partitions per $N$ (Figure~\ref{fig:scaling}), RGV is above
% DeepConf at every $N$; the gap grows monotonically from $+3.4\%$ at
% $N{=}2$ to $+5.4\%$ at $N{=}8$. RGV at $N{=}4$ ($65.9\%$) already
% matches DeepConf at $N{=}8$ ($65.7\%$), a $50\%$ rollout-budget
% saving at iso-accuracy. The advantage persists across all $N$ and grows with the rollout
% budget, the regime where test-time scaling
% \citep{snell2024scalecompute,brown2024monkeys} is most relevant.

Sub-sampling the eight rollouts to $N \in \{1,\dots,8\}$ at $60$
random partitions per $N$ (Figure~\ref{fig:scaling}), RGV at
$N{=}4$ ($65.9\%$) already matches DeepConf at $N{=}8$ ($65.7\%$):
the same accuracy at half the rollout cost. The gap grows
monotonically from $+3.4\%$ at $N{=}2$ to $+5.4\%$ at $N{=}8$,
meaning RGV benefits more, not less, from additional rollouts. This
is the regime where test-time scaling
\citep{snell2024scalecompute,brown2024monkeys} is most relevant,
and where a better voting rule yields compounding returns.

\paragraph{Cost.}
% RGV uses only the emitted prose and the retrieved snippets already
% in the trajectory: no extra call, no logprobs, $\sim$0.3\,ms per
% rollout on one CPU thread. By contrast, DeepConf requires per-token
% logprobs, and aggregator methods
% \citep{lee2026aggagent,chen2023universalsc} add at least one
% rollout-equivalent of inference. When combined with the
% iso-accuracy scaling above, RGV offers a favourable cost--accuracy
% trade-off.

RGV uses only the emitted prose and the retrieved snippets already
in the trajectory: no extra call, no logprobs, $\sim$0.3\,ms per
rollout on one CPU thread. By contrast, DeepConf requires per-token
logprobs, and aggregator methods
\citep{lee2026aggagent,chen2023universalsc} add at least one
rollout-equivalent of inference. When combined with the
iso-accuracy scaling above, RGV offers a favourable cost--accuracy
trade-off.

\paragraph{Max-over-docs ablation.}

\begin{table}[t]
  \centering\small
  \begin{tabular}{lc}
    \toprule
    \textbf{Reduction} & \textbf{Accuracy (\%)} \\
    \midrule
    Simple Majority (no RGV) & 62.0 \\
    \midrule
    $\min_d$           & 69.3 \\
    $\mathrm{mean}_d$  & 70.5 \\
    $\max_d - \min_d$  & 70.5 \\
    $\max_d$ (headline RGV)  & \textbf{71.1} \\
    \bottomrule
  \end{tabular}
  \caption{\textbf{Doc-set reduction ablation} ($N{=}8$, BrowseComp-Plus, Tongyi-DeepResearch).}
  \label{tab:reduction}
\end{table}

% \S\ref{sec:method} argues for $\max$ over the retrieved-document set
% to avoid dilution from irrelevant documents.
% Table~\ref{tab:reduction} validates this: $\max$ consistently leads,
% though all four reductions beat simple majority by $+7$ to $+9\%$.
% The takeaway is that the \emph{signal source} (prose-vs-retrieval
% overlap read outside the contaminated context) matters more than the
% choice of reduction; $\max$ adds a further $+0.6$ to $+1.8\%$ by
% focusing on the single most relevant document.

\S\ref{sec:method} argues for $\max$ over the retrieved-document set
to avoid dilution from irrelevant documents.
Table~\ref{tab:reduction} validates this: $\max$ consistently leads,
though all four reductions beat simple majority by $+7$ to $+9\%$.
The fact that even $\min$ (the worst-matching document) still yields
a $+7.3\%$ gain over simple majority is revealing: the bulk of
RGV's advantage comes from the \emph{signal source} itself, i.e.\
reading prose-vs-retrieval overlap outside the contaminated context.
The choice of reduction is secondary; $\max$ adds a further $+0.6$
to $+1.8\%$ by focusing on the single most relevant document and
avoiding dilution from irrelevant retrievals.

% ======================================================================
\subsection{Error analysis}
\label{sec:exp-error}

RGV selects the correct answer on $590$ of $616$ questions where a
correct rollout exists ($95.8\%$). The $26$ failures ($4.2\%$)
follow a well-characterised pattern: in $24$ cases a wrong rollout
retrieves documents about the right \emph{topic} but draws the
wrong \emph{conclusion}, receiving a higher grounding score because
it echoes topically relevant---but factually incorrect---evidence.
We call this \emph{well-grounded but wrong}. This residual is small relative to the $+5.4\%$ overall gain; it
concentrates on high-diversity questions (mean $4.9$ unique clusters
vs.\ $4.0$) where many plausible answers compete.
Appendix~\ref{app:error-cases} contrasts three representative
successes with three failure cases.
\section{Related work}
\label{sec:related}

\paragraph{Confidence-based aggregation for single-turn reasoning.}
Sample-and-aggregate test-time compute
\citep{wang2022selfconsistency,snell2024scalecompute,brown2024monkeys,
wang2024mixtureofagents} typically weights the vote by a confidence
proxy: self-reported
confidence~\citep{wang2024softsc,taubenfeld2025cisc}, sliding-window
token logprobs~\citep{fu2025deepconf}, an extra LLM
judge~\citep{chen2023universalsc,thirukovalluru2024atomicsc}, or
saturation-based budget~\citep{aggarwal2023letssample}. These
assume the question is the only conditioning, with no tool outputs in
context.

\paragraph{The breakdown in multi-turn tool-using agents.}
Within the broader class of tool-using agents, we focus on search
agents, the subclass where retrieved documents dominate the context.
Recent work reports that confidence transfers poorly to this setting,
but examines a different signal class.
\citet{xuan2026dichotomy} document a \emph{confidence dichotomy}---%
evidence tools induce overconfidence in \emph{verbalised}
confidence and recalibrate it via RL fine-tuning;
BrowseConf~\citep{ou2025browseconf} uses verbalised confidence as
a retry trigger. Both target the verbalised channel rather than the
token-logprob signal used by logit-based voters, and neither
operates at the voting layer. \citet{wang2024softsc} soften
majority voting with token logprobs on interactive tasks without
retrieval. Broader miscalibration accounts
\citep{kadavath2022knowwhattheyknow,tian2023justaskcalibrated,
xiong2023canllmexpress,kuhn2023semanticuncertainty} share our
framing; our contribution is to address the mechanism at the voting
layer without retraining or a second model.

\paragraph{Trajectory-aware aggregators.}
A complementary thread adds an extra model: AggAgent
\citep{lee2026aggagent} aggregates parallel trajectories;
ParallelMuse \citep{li2025parallelmuse} compresses partial ones for
an aggregator; PRM-style verifiers
\citep{cobbe2021verifiers,lightman2024prm} and self-correction
agents \citep{shinn2023reflexion,zhou2024selfdiscover} score
intermediate steps. These help but add at least one
rollout-equivalent of inference; RGV adds none.

\paragraph{Faithfulness and grounding metrics.}
RGV's lexical-overlap signal is a voting-layer use of a familiar
intuition from summarisation and RAG evaluation
\citep{maynez2020faithfulness,lin2004rouge,
min2023factscore,es2024ragas,gao2023ragsurvey}. The primitive itself
is long-standing: \citet{grusky-etal-2018-newsroom} measure extractive
\emph{coverage} as the fraction of \emph{summary} tokens drawn from the
source article, the same answer-side normalisation we adopt in
\S\ref{sec:method}, and
\citet{shuster-etal-2021-retrieval-augmentation} use unigram overlap
between a response and its grounding knowledge as a hallucination
proxy, with a Rare-F1 variant that discounts common words for exactly
the reason Figure~\ref{fig:why}(\subref{fig:why-a}) finds them
uninformative. Those metrics are evaluated post-hoc against a single
output. Our contribution is the \emph{layer} at which the primitive is
applied, weighted voting over parallel rollouts, and the reason it
has to be read outside the model at all: copy-inflation
(\S\ref{sec:motivation}) is what breaks the internal alternative.

% ======================================================================

\section{Conclusion}
We identified copy-inflation, the mechanism by which retrieved documents inflate logprob-based confidence in search agents, and proposed RGV, which reads the voting signal from \emph{outside the contaminated context}. Across four benchmarks and five LLMs, RGV outperforms both simple majority and confidence-based voting while adding minimal cost. Our findings suggest that when retrieved evidence contaminates the model context, aggregation signals drawn from environment-derived records offer a more reliable foundation than internal model signals. The diagnosis is not specific to voting: any mechanism that consumes an agent's token-level confidence, whether for early stopping, routing, abstention, confidence-shaped rewards, or logprob-based hallucination detection, reads the same contaminated signal and inherits the same failure. We believe this direction may extend beyond voting to other settings where context and evidence are no longer separable.
\section*{Limitations}

\paragraph{RGV inherits retrieval quality.}
The score is read from the retrieval log, so it can be no better than what
retrieval returned. Appendix~\ref{app:weak-retrieval} maps this boundary with
four cells. Where a weaker retriever genuinely degrades the agent (BM25 in
place of the dense retriever), the grounded vote keeps its margin; where the
weaker stack still suffices for the benchmark, all aggregation rules converge
and RGV neither helps nor hurts; under outright corpus mismatch it is the only
rule that stays above the single-rollout average, though its own margin is
thin. Degradation is graceful, and the copy fraction of a cell predicts which
regime applies. But on a task where retrieval contributes little, so does RGV.

\paragraph{Grounded is not correct, and short answers weaken the signal.}
RGV estimates retrieval success, not truth. Section~\ref{sec:exp-error}
measures the residual: on $26$ of $616$ questions ($4.2\%$) a wrong rollout
retrieves documents on the right topic and outscores a correct one. The signal
is surface-level overlap with no entailment check, so this is intrinsic to the
design rather than a tuning artefact. Relatedly, because the score normalises
by the answer prose, it weakens as that prose becomes very short: forcing
single-entity answers reduces it to answer containment and the vote merely
matches DeepConf (Appendix~\ref{app:prompt-format}).

\paragraph{Scope.}
We study multi-turn search agents, where retrieved documents are the primary
evidence source; agents relying on non-retrieval tools such as code
interpreters would need a different grounding signal, and on
reasoning-intensive tasks the gains shrink (Appendix~\ref{app:hle}). All
experiments are in English. Finally, RGV trusts the retrieval log, so an
adversarially poisoned corpus would raise the score of rollouts that copy from
it. This vulnerability is shared by any method that reads that log. Accidental noise is
not a problem (injecting up to $16$ irrelevant documents per rollout leaves the
vote unchanged), but adversarial robustness remains open.

% ======================================================================

\section*{Acknowledgments}
This work was partially supported by Google Research Award, Google ML \& System Junior Faculty
Award, Amazon Research Award, Fireworks AI, Intel, Li Auto, Moffett AI, and CMU CyLab Seed
funding. This material is also based upon work
supported by the National Science Foundation under Grant No. 2504353. Any opinions, findings,
and conclusions or recommendations expressed are
those of the authors and do not necessarily reflect
the views of the National Science Foundation.
This research is based upon work supported in
part by the Office of the Director of National Intelligence (ODNI), Intelligence Advanced Research
Projects Activity (IARPA), via 560000C260017.
The views and conclusions contained herein are
those of the authors and should not be interpreted
as necessarily representing the official policies, either expressed or implied, of ODNI, IARPA, or
the U.S. Government. The U.S. Government is
authorized to reproduce and distribute reprints for
governmental purposes notwithstanding any copyright annotation therein.

This work was supported by Institute of Information \& communications Technology Planning \& Evaluation (IITP) grant funded by the Korea government(MSIT) (RS-2022-00143911, AI Excellence Global Innovative Leader Education Program)

% Custom bibliography entries only
\bibliography{refs}

\begin{thebibliography}{51}
\providecommand{\natexlab}[1]{#1}

\bibitem[{Aggarwal et~al.(2023)Aggarwal, Madaan, Yang, and
  Mausam}]{aggarwal2023letssample}
Pranjal Aggarwal, Aman Madaan, Yiming Yang, and Mausam. 2023.
\newblock \href {https://arxiv.org/abs/2305.11860} {Let's sample step by step:
  Adaptive-consistency for efficient reasoning and coding with llms}.
\newblock \emph{Preprint}, arXiv:2305.11860.

\bibitem[{Brown et~al.(2024)Brown, Juravsky, Ehrlich, Clark, Le, Ré, and
  Mirhoseini}]{brown2024monkeys}
Bradley Brown, Jordan Juravsky, Ryan Ehrlich, Ronald Clark, Quoc~V. Le,
  Christopher Ré, and Azalia Mirhoseini. 2024.
\newblock \href {https://arxiv.org/abs/2407.21787} {Large language monkeys:
  Scaling inference compute with repeated sampling}.
\newblock \emph{Preprint}, arXiv:2407.21787.

\bibitem[{Chen et~al.(2023)Chen, Aksitov, Alon, Ren, Xiao, Yin, Prakash,
  Sutton, Wang, and Zhou}]{chen2023universalsc}
Xinyun Chen, Renat Aksitov, Uri Alon, Jie Ren, Kefan Xiao, Pengcheng Yin,
  Sushant Prakash, Charles Sutton, Xuezhi Wang, and Denny Zhou. 2023.
\newblock \href {https://arxiv.org/abs/2311.17311} {Universal self-consistency
  for large language model generation}.
\newblock \emph{Preprint}, arXiv:2311.17311.

\bibitem[{Chen et~al.(2025)Chen, Ma, Zhuang, Nie, Zou, Liu, Green, Patel, Meng,
  Su, Sharifymoghaddam, Li, Hong, Shi, Liu, Thakur, Zhang, Gao, Chen, and
  Lin}]{chen2025browsecompplus}
Zijian Chen, Xueguang Ma, Shengyao Zhuang, Ping Nie, Kai Zou, Andrew Liu,
  Joshua Green, Kshama Patel, Ruoxi Meng, Mingyi Su, Sahel Sharifymoghaddam,
  Yanxi Li, Haoran Hong, Xinyu Shi, Xuye Liu, Nandan Thakur, Crystina Zhang,
  Luyu Gao, Wenhu Chen, and Jimmy Lin. 2025.
\newblock \href {https://arxiv.org/abs/2508.06600} {Browsecomp-plus: A more
  fair and transparent evaluation benchmark of deep-research agent}.
\newblock \emph{Preprint}, arXiv:2508.06600.

\bibitem[{Cobbe et~al.(2021)Cobbe, Kosaraju, Bavarian, Chen, Jun, Kaiser,
  Plappert, Tworek, Hilton, Nakano, Hesse, and Schulman}]{cobbe2021verifiers}
Karl Cobbe, Vineet Kosaraju, Mohammad Bavarian, Mark Chen, Heewoo Jun, Lukasz
  Kaiser, Matthias Plappert, Jerry Tworek, Jacob Hilton, Reiichiro Nakano,
  Christopher Hesse, and John Schulman. 2021.
\newblock \href {https://arxiv.org/abs/2110.14168} {Training verifiers to solve
  math word problems}.
\newblock \emph{Preprint}, arXiv:2110.14168.

\bibitem[{Es et~al.(2024)Es, James, Espinosa~Anke, and
  Schockaert}]{es2024ragas}
Shahul Es, Jithin James, Luis Espinosa~Anke, and Steven Schockaert. 2024.
\newblock \href {https://doi.org/10.18653/v1/2024.eacl-demo.16} {{RAGA}s:
  Automated evaluation of retrieval augmented generation}.
\newblock In \emph{Proceedings of the 18th Conference of the European Chapter
  of the Association for Computational Linguistics: System Demonstrations},
  pages 150--158, St. Julians, Malta. Association for Computational
  Linguistics.

\bibitem[{Fawcett(2006)}]{fawcett2006roc}
Tom Fawcett. 2006.
\newblock \href {https://doi.org/10.1016/j.patrec.2005.10.010} {An introduction
  to roc analysis}.
\newblock \emph{Pattern Recognition Letters}, 27(8):861--874.
\newblock ROC Analysis in Pattern Recognition.

\bibitem[{Fu et~al.(2025)Fu, Wang, Tian, and Zhao}]{fu2025deepconf}
Yichao Fu, Xuewei Wang, Yuandong Tian, and Jiawei Zhao. 2025.
\newblock \href {https://arxiv.org/abs/2508.15260} {Deep think with
  confidence}.
\newblock \emph{Preprint}, arXiv:2508.15260.

\bibitem[{Gao et~al.(2024)Gao, Xiong, Gao, Jia, Pan, Bi, Dai, Sun, Wang, and
  Wang}]{gao2023ragsurvey}
Yunfan Gao, Yun Xiong, Xinyu Gao, Kangxiang Jia, Jinliu Pan, Yuxi Bi, Yi~Dai,
  Jiawei Sun, Meng Wang, and Haofen Wang. 2024.
\newblock \href {https://arxiv.org/abs/2312.10997} {Retrieval-augmented
  generation for large language models: A survey}.
\newblock \emph{Preprint}, arXiv:2312.10997.

\bibitem[{GLM-5-Team et~al.(2026)GLM-5-Team, :, Zeng, Lv, Hou, Du, Zheng, Chen,
  Yin, Ge, Huang, Xie, Zhu, Yin, Wang, Pan, Zeng, Zhang, Wang, Chen, Zhang,
  Jiao, Guo, Wang, Du, Wu, Wang, Li, Fan, Zhong, Liu, Zhao, Du, Dong, Lu,
  Shuang-Li, Cao, Liu, Jiang, Chen, Zhang, Huang, Dong, Xu, Wei, An, Niu, Zhu,
  Wen, Cen, Bai, Qiao, Wang, Wang, Zhu, Liu, Li, Wang, Wen, Huang, Cai, Yu, Li,
  Hu, Zhang, Zhang, Lin, Yang, Wang, Ai, Zhu, Yi, Chen, Wen, Sun, Zhao, Hu,
  Zhang, Liu, Zhang, Peng, Tai, Zhang, Liu, Wang, Yan, Ge, Liu, Chu, Zhao,
  Wang, Zhao, Ren, Wang, Zhang, Gui, Zhao, Li, An, Li, Yuan, Du, Liu, Zhi,
  Duan, Zhou, Wei, Wang, Luo, Zhang, Sha, Xu, Wu, Ding, Chen, Li, Lin, Ta, Zou,
  Song, Yang, Tu, Yang, Wu, Zhang, Li, Li, Fan, Qin, Tian, Zhang, Yu, Liang,
  Kuang, Cheng, Li, Yan, Hu, Ling, Fan, Xia, Zhang, Zhang, Pan, Zou, Zhang,
  Liu, Wu, Li, Wang, Zhu, Tan, Zhou, Pan, Zhang, Su, Geng, Yan, Tan, Bi, Shen,
  Yang, Li, Liu, Wang, Li, Wu, Zhang, Duan, Zhang, Liu, Jiang, Yan, Zhang, Wei,
  Chen, Feng, Yao, Chai, Wang, Zhang, Xu, Huang, Wang, Li, Dong, and
  Tang}]{zai2026glm5}
GLM-5-Team, :, Aohan Zeng, Xin Lv, Zhenyu Hou, Zhengxiao Du, Qinkai Zheng, Bin
  Chen, Da~Yin, Chendi Ge, Chenghua Huang, Chengxing Xie, Chenzheng Zhu,
  Congfeng Yin, Cunxiang Wang, Gengzheng Pan, Hao Zeng, Haoke Zhang, Haoran
  Wang, and 168 others. 2026.
\newblock \href {https://arxiv.org/abs/2602.15763} {Glm-5: from vibe coding to
  agentic engineering}.
\newblock \emph{Preprint}, arXiv:2602.15763.

\bibitem[{Grusky et~al.(2018)Grusky, Naaman, and
  Artzi}]{grusky-etal-2018-newsroom}
Max Grusky, Mor Naaman, and Yoav Artzi. 2018.
\newblock \href {https://doi.org/10.18653/v1/N18-1065} {{N}ewsroom: A dataset
  of 1.3 million summaries with diverse extractive strategies}.
\newblock In \emph{Proceedings of the 2018 Conference of the North {A}merican
  Chapter of the Association for Computational Linguistics: Human Language
  Technologies, Volume 1 (Long Papers)}, pages 708--719, New Orleans,
  Louisiana. Association for Computational Linguistics.

\bibitem[{Jaccard(1912)}]{jaccard1912index}
Paul Jaccard. 1912.
\newblock \href {http://www.jstor.org/stable/2427226} {The distribution of the
  flora in the alpine zone}.
\newblock \emph{The New Phytologist}, 11(2):37--50.

\bibitem[{Jin et~al.(2025)Jin, Zeng, Yue, Yoon, Arik, Wang, Zamani, and
  Han}]{jin2025searchr1}
Bowen Jin, Hansi Zeng, Zhenrui Yue, Jinsung Yoon, Sercan Arik, Dong Wang, Hamed
  Zamani, and Jiawei Han. 2025.
\newblock Search-r1: Training llms to reason and leverage search engines with
  reinforcement learning.
\newblock \emph{arXiv preprint arXiv:2503.09516}.

\bibitem[{Kadavath et~al.(2022)Kadavath, Conerly, Askell, Henighan, Drain,
  Perez, Schiefer, Hatfield-Dodds, DasSarma, Tran-Johnson, Johnston, El-Showk,
  Jones, Elhage, Hume, Chen, Bai, Bowman, Fort, Ganguli, Hernandez, Jacobson,
  Kernion, Kravec, Lovitt, Ndousse, Olsson, Ringer, Amodei, Brown, Clark,
  Joseph, Mann, McCandlish, Olah, and Kaplan}]{kadavath2022knowwhattheyknow}
Saurav Kadavath, Tom Conerly, Amanda Askell, Tom Henighan, Dawn Drain, Ethan
  Perez, Nicholas Schiefer, Zac Hatfield-Dodds, Nova DasSarma, Eli
  Tran-Johnson, Scott Johnston, Sheer El-Showk, Andy Jones, Nelson Elhage,
  Tristan Hume, Anna Chen, Yuntao Bai, Sam Bowman, Stanislav Fort, and 17
  others. 2022.
\newblock \href {https://arxiv.org/abs/2207.05221} {Language models (mostly)
  know what they know}.
\newblock \emph{Preprint}, arXiv:2207.05221.

\bibitem[{Krishna et~al.(2025)Krishna, Krishna, Mohananey, Schwarcz, Stambler,
  Upadhyay, and Faruqui}]{krishna2024frames}
Satyapriya Krishna, Kalpesh Krishna, Anhad Mohananey, Steven Schwarcz, Adam
  Stambler, Shyam Upadhyay, and Manaal Faruqui. 2025.
\newblock Fact, fetch, and reason: A unified evaluation of retrieval-augmented
  generation.
\newblock In \emph{Proceedings of the 2025 Conference of the Nations of the
  Americas Chapter of the Association for Computational Linguistics: Human
  Language Technologies (Volume 1: Long Papers)}, pages 4745--4759.

\bibitem[{Kuhn et~al.(2023)Kuhn, Gal, and
  Farquhar}]{kuhn2023semanticuncertainty}
Lorenz Kuhn, Yarin Gal, and Sebastian Farquhar. 2023.
\newblock Semantic uncertainty: Linguistic invariances for uncertainty
  estimation in natural language generation.
\newblock \emph{arXiv preprint arXiv:2302.09664}.

\bibitem[{Lee et~al.(2026)Lee, Yen, Ye, and Chen}]{lee2026aggagent}
Yoonsang Lee, Howard Yen, Xi~Ye, and Danqi Chen. 2026.
\newblock \href {https://arxiv.org/abs/2604.11753} {Agentic aggregation for
  parallel scaling of long-horizon agentic tasks}.
\newblock \emph{Preprint}, arXiv:2604.11753.

\bibitem[{Li et~al.(2025)Li, Zhang, Wu, Yin, Tao, Zhao, Zhang, Shen, Fang, Xie,
  Zhou, and Jiang}]{li2025parallelmuse}
Baixuan Li, Dingchu Zhang, Jialong Wu, Wenbiao Yin, Zhengwei Tao, Yida Zhao,
  Liwen Zhang, Haiyang Shen, Runnan Fang, Pengjun Xie, Jingren Zhou, and Yong
  Jiang. 2025.
\newblock \href {https://arxiv.org/abs/2510.24698} {Parallelmuse: Agentic
  parallel thinking for deep information seeking}.
\newblock \emph{Preprint}, arXiv:2510.24698.

\bibitem[{Lightman et~al.(2023)Lightman, Kosaraju, Burda, Edwards, Baker, Lee,
  Leike, Schulman, Sutskever, and Cobbe}]{lightman2024prm}
Hunter Lightman, Vineet Kosaraju, Yura Burda, Harri Edwards, Bowen Baker, Teddy
  Lee, Jan Leike, John Schulman, Ilya Sutskever, and Karl Cobbe. 2023.
\newblock \href {https://arxiv.org/abs/2305.20050} {Let's verify step by step}.
\newblock \emph{Preprint}, arXiv:2305.20050.

\bibitem[{Lin(2004)}]{lin2004rouge}
Chin-Yew Lin. 2004.
\newblock \href {https://aclanthology.org/W04-1013/} {{ROUGE}: A package for
  automatic evaluation of summaries}.
\newblock In \emph{Text Summarization Branches Out}, pages 74--81, Barcelona,
  Spain. Association for Computational Linguistics.

\bibitem[{Luan et~al.(2021)Luan, Eisenstein, Toutanova, and
  Collins}]{luan-etal-2021-sparse}
Yi~Luan, Jacob Eisenstein, Kristina Toutanova, and Michael Collins. 2021.
\newblock \href {https://doi.org/10.1162/tacl_a_00369} {Sparse, dense, and
  attentional representations for text retrieval}.
\newblock \emph{Transactions of the Association for Computational Linguistics},
  9:329--345.

\bibitem[{Manakul et~al.(2023)Manakul, Liusie, and
  Gales}]{manakul-etal-2023-selfcheckgpt}
Potsawee Manakul, Adian Liusie, and Mark Gales. 2023.
\newblock \href {https://doi.org/10.18653/v1/2023.emnlp-main.557}
  {{S}elf{C}heck{GPT}: Zero-resource black-box hallucination detection for
  generative large language models}.
\newblock In \emph{Proceedings of the 2023 Conference on Empirical Methods in
  Natural Language Processing}, pages 9004--9017, Singapore. Association for
  Computational Linguistics.

\bibitem[{Maynez et~al.(2020)Maynez, Narayan, Bohnet, and
  McDonald}]{maynez2020faithfulness}
Joshua Maynez, Shashi Narayan, Bernd Bohnet, and Ryan McDonald. 2020.
\newblock On faithfulness and factuality in abstractive summarization.
\newblock In \emph{Proceedings of the 58th annual meeting of the association
  for computational linguistics}, pages 1906--1919.

\bibitem[{Mialon et~al.(2023)Mialon, Fourrier, Swift, Wolf, LeCun, and
  Scialom}]{mialon2023gaia}
Grégoire Mialon, Clémentine Fourrier, Craig Swift, Thomas Wolf, Yann LeCun,
  and Thomas Scialom. 2023.
\newblock \href {https://arxiv.org/abs/2311.12983} {Gaia: a benchmark for
  general ai assistants}.
\newblock \emph{Preprint}, arXiv:2311.12983.

\bibitem[{Min et~al.(2023)Min, Krishna, Lyu, Lewis, tau Yih, Koh, Iyyer,
  Zettlemoyer, and Hajishirzi}]{min2023factscore}
Sewon Min, Kalpesh Krishna, Xinxi Lyu, Mike Lewis, Wen tau Yih, Pang~Wei Koh,
  Mohit Iyyer, Luke Zettlemoyer, and Hannaneh Hajishirzi. 2023.
\newblock \href {https://arxiv.org/abs/2305.14251} {Factscore: Fine-grained
  atomic evaluation of factual precision in long form text generation}.
\newblock \emph{Preprint}, arXiv:2305.14251.

\bibitem[{MiniMax et~al.(2025)MiniMax, :, Chen, Li, Gong, Jiang, Fei, Yang,
  Shan, Yu, Wang, Zhu, Xiao, Du, Zhang, Qiao, Zhang, Du, Guo, Chen, Ding, Sun,
  Li, Jiao, Zhou, Zhang, Ding, Sun, Feng, Cai, Zhu, Sun, Zhuang, Cai, Song,
  Zhu, Li, Tian, Liu, Xu, Yan, Liu, He, Feng, Yang, Xiao, Han, Wang, Yu, Feng,
  Li, Zheng, Du, Yang, Zeng, Yu, Tao, Chi, Zhang, Lin, Hu, Di, Gao, Li, Zhao,
  Ren, Xu, Li, Wang, Tian, Leng, Chen, Chen, Shi, Weng, Guan, Yu, Li, Zhu, Li,
  Cai, Liang, Cheng, Kong, Li, Chen, Song, Luo, Su, Li, Han, Hou, Lu, Zou,
  Shen, Gong, Ma, Wang, Shi, Zhong, Duan, Fu, Hu, Gao, Fan, Yang, Li, Hu,
  Huang, Li, Xu, Mao, Shi, Wenren, Li, Li, Tian, Zhu, Fan, Wu, Xu, Yu, Lyu,
  Jiang, Gao, Wu, Song, and Sun}]{minimax2025m1}
MiniMax, :, Aili Chen, Aonian Li, Bangwei Gong, Binyang Jiang, Bo~Fei, Bo~Yang,
  Boji Shan, Changqing Yu, Chao Wang, Cheng Zhu, Chengjun Xiao, Chengyu Du, Chi
  Zhang, Chu Qiao, Chunhao Zhang, Chunhui Du, Congchao Guo, and 109 others.
  2025.
\newblock \href {https://arxiv.org/abs/2506.13585} {Minimax-m1: Scaling
  test-time compute efficiently with lightning attention}.
\newblock \emph{Preprint}, arXiv:2506.13585.

\bibitem[{Nakano et~al.(2021)Nakano, Hilton, Balaji, Wu, Ouyang, Kim, Hesse,
  Jain, Kosaraju, Saunders et~al.}]{nakano2021webgpt}
Reiichiro Nakano, Jacob Hilton, Suchir Balaji, Jeff Wu, Long Ouyang, Christina
  Kim, Christopher Hesse, Shantanu Jain, Vineet Kosaraju, William Saunders, and
  1 others. 2021.
\newblock Webgpt: Browser-assisted question-answering with human feedback.
\newblock \emph{arXiv preprint arXiv:2112.09332}.

\bibitem[{OpenAI et~al.(2025)OpenAI, :, Agarwal, Ahmad, Ai, Altman, Applebaum,
  Arbus, Arora, Bai, Baker, Bao, Barak, Bennett, Bertao, Brett, Brevdo,
  Brockman, Bubeck, Chang, Chen, Chen, Cheung, Clark, Cook, Dukhan, Dvorak,
  Fives, Fomenko, Garipov, Georgiev, Glaese, Gogineni, Goucher, Gross, Guzman,
  Hallman, Hehir, Heidecke, Helyar, Hu, Huet, Huh, Jain, Johnson, Koch, Kofman,
  Kundel, Kwon, Kyrylov, Le, Leclerc, Lennon, Lessans, Lezcano-Casado, Li, Li,
  Lin, Liss, Lily, Liu, Liu, Lu, Lu, Martinovic, McCallum, McGrath, McKinney,
  McLaughlin, Mei, Mostovoy, Mu, Myles, Neitz, Nichol, Pachocki, Paino, Palmie,
  Pantuliano, Parascandolo, Park, Pathak, Paz, Peran, Pimenov, Pokrass, Proehl,
  Qiu, Raila, Raso, Ren, Richardson, Robinson, Rotsted, Salman, Sanjeev,
  Schwarzer, Sculley, Sikchi, Simon, Singhal, Song, Stuckey, Sun, Tillet,
  Toizer, Tsimpourlas, Vyas, Wallace, Wang, Wang, Watkins, Weil, Wendling,
  Whinnery, Whitney, Wong, Yang, Yang, Yasunaga, Ying, Zaremba, Zhan, Zhang,
  Zhang, Zhang, and Zhao}]{openai2025gptoss}
OpenAI, :, Sandhini Agarwal, Lama Ahmad, Jason Ai, Sam Altman, Andy Applebaum,
  Edwin Arbus, Rahul~K. Arora, Yu~Bai, Bowen Baker, Haiming Bao, Boaz Barak,
  Ally Bennett, Tyler Bertao, Nivedita Brett, Eugene Brevdo, Greg Brockman,
  Sebastien Bubeck, and 108 others. 2025.
\newblock \href {https://arxiv.org/abs/2508.10925} {gpt-oss-120b \& gpt-oss-20b
  model card}.
\newblock \emph{Preprint}, arXiv:2508.10925.

\bibitem[{Ou et~al.(2025)Ou, Li, Yin, Zhang, Zhang, Wu, Ye, Qiao, Xie, Zhou,
  and Jiang}]{ou2025browseconf}
Litu Ou, Kuan Li, Huifeng Yin, Liwen Zhang, Zhongwang Zhang, Xixi Wu, Rui Ye,
  Zile Qiao, Pengjun Xie, Jingren Zhou, and Yong Jiang. 2025.
\newblock \href {https://arxiv.org/abs/2510.23458} {Browseconf:
  Confidence-guided test-time scaling for web agents}.
\newblock \emph{Preprint}, arXiv:2510.23458.

\bibitem[{Phan et~al.(2025)Phan, Gatti, Han, Li, Hu, Zhang, Zhang, Shaaban,
  Ling, Shi et~al.}]{phan2025hle}
Long Phan, Alice Gatti, Ziwen Han, Nathaniel Li, Josephina Hu, Hugh Zhang, Chen
  Bo~Calvin Zhang, Mohamed Shaaban, John Ling, Sean Shi, and 1 others. 2025.
\newblock Humanity's last exam.
\newblock \emph{arXiv preprint arXiv:2501.14249}.

\bibitem[{Robertson and Zaragoza(2009)}]{robertson2009bm25}
Stephen Robertson and Hugo Zaragoza. 2009.
\newblock \href {https://doi.org/10.1561/1500000019} {The probabilistic
  relevance framework: Bm25 and beyond}.
\newblock \emph{Found. Trends Inf. Retr.}, 3(4):333–389.

\bibitem[{Schick et~al.(2023)Schick, Dwivedi-Yu, Dess\'{\i}, Raileanu, Lomeli,
  Hambro, Zettlemoyer, Cancedda, and Scialom}]{schick2023toolformer}
Timo Schick, Jane Dwivedi-Yu, Roberto Dess\'{\i}, Roberta Raileanu, Maria
  Lomeli, Eric Hambro, Luke Zettlemoyer, Nicola Cancedda, and Thomas Scialom.
  2023.
\newblock Toolformer: language models can teach themselves to use tools.
\newblock In \emph{Proceedings of the 37th International Conference on Neural
  Information Processing Systems}, NIPS '23, Red Hook, NY, USA. Curran
  Associates Inc.

\bibitem[{Sciavolino et~al.(2021)Sciavolino, Zhong, Lee, and
  Chen}]{sciavolino-etal-2021-simple}
Christopher Sciavolino, Zexuan Zhong, Jinhyuk Lee, and Danqi Chen. 2021.
\newblock \href {https://doi.org/10.18653/v1/2021.emnlp-main.496} {Simple
  entity-centric questions challenge dense retrievers}.
\newblock In \emph{Proceedings of the 2021 Conference on Empirical Methods in
  Natural Language Processing}, pages 6138--6148, Online and Punta Cana,
  Dominican Republic. Association for Computational Linguistics.

\bibitem[{Shinn et~al.(2023)Shinn, Cassano, Gopinath, Narasimhan, and
  Yao}]{shinn2023reflexion}
Noah Shinn, Federico Cassano, Ashwin Gopinath, Karthik Narasimhan, and Shunyu
  Yao. 2023.
\newblock Reflexion: Language agents with verbal reinforcement learning.
\newblock \emph{Advances in neural information processing systems},
  36:8634--8652.

\bibitem[{Shuster et~al.(2021)Shuster, Poff, Chen, Kiela, and
  Weston}]{shuster-etal-2021-retrieval-augmentation}
Kurt Shuster, Spencer Poff, Moya Chen, Douwe Kiela, and Jason Weston. 2021.
\newblock \href {https://doi.org/10.18653/v1/2021.findings-emnlp.320}
  {Retrieval augmentation reduces hallucination in conversation}.
\newblock In \emph{Findings of the Association for Computational Linguistics:
  EMNLP 2021}, pages 3784--3803, Punta Cana, Dominican Republic. Association
  for Computational Linguistics.

\bibitem[{Snell et~al.(2024)Snell, Lee, Xu, and Kumar}]{snell2024scalecompute}
Charlie Snell, Jaehoon Lee, Kelvin Xu, and Aviral Kumar. 2024.
\newblock \href {https://arxiv.org/abs/2408.03314} {Scaling llm test-time
  compute optimally can be more effective than scaling model parameters}.
\newblock \emph{Preprint}, arXiv:2408.03314.

\bibitem[{Taubenfeld et~al.(2025)Taubenfeld, Sheffer, Ofek, Feder, Goldstein,
  Gekhman, and Yona}]{taubenfeld2025cisc}
Amir Taubenfeld, Tom Sheffer, Eran Ofek, Amir Feder, Ariel Goldstein, Zorik
  Gekhman, and Gal Yona. 2025.
\newblock Confidence improves self-consistency in llms.
\newblock In \emph{Findings of the Association for Computational Linguistics:
  ACL 2025}, pages 20090--20111.

\bibitem[{Team et~al.(2025{\natexlab{a}})Team, Bai, Bao, Charles, Chen, Chen,
  Chen, Chen, Chen, Chen, Chen, Chen, Chen, Chen, Chen, Cui, Ding, Dong, Du,
  Du, Du, Du, Fan, Feng, Fu, Gao, Gao, Gao, Gao, Gao, Ge, Geng, Gu, Gu, Guan,
  Guo, Guo, Hao, He, He, He, He, Hong, Hu, Hu, Hu, Huang, Huang, Huang, Jiang,
  Jiang, Jin, Kang, Lai, Li, Li, Li, Li, Li, Li, Li, Li, Li, Li, Lin, Lin, Lin,
  Liu, Liu, Liu, Liu, Liu, Liu, Liu, Liu, Liu, Liu, Liu, Liu, Liu, Liu, Liu,
  Lu, Lu, Lu, Luo, Ma, Ma, Ma, Mao, Mei, Men, Miao, Pan, Peng, Qin, Qin, Qu,
  Shang, Shi, Shi, Song, Su, Su, Sui, Sun, Sung, Tai, Tang, Tao, Teng, Tian,
  Wang, Wang, Wang, Wang, Wang, Wang, Wang, Wang, Wang, Wang, Wang, Wang, Wang,
  Wang, Wang, Wang, Wang, Wang, Wang, Wang, Wang, Wei, Wei, Wu, Wu, Wu, Wu,
  Xiao, Xie, Xie, Xiong, Xu, Xu, Xu, Xu, Xu, Xu, Xu, Xu, Xu, Xu, Xu, Yan, Yan,
  Yang, Yang, Yang, Yang, Yang, Yang, Yang, Yao, Yao, Ye, Ye, Yin, Yu, Yuan,
  Yuan, Yuan, Yuan, Zhan, Zhang, Zhang, Zhang, Zhang, Zhang, Zhang, Zhang,
  Zhang, Zhang, Zhang, Zhang, Zhang, Zhang, Zhao, Zhao, Zhao, Zheng, Zheng,
  Zhong, Zhou, Zhou, Zhou, Zhu, Zhu, Zhuang, and Zu}]{moonshot2025kimik2}
Kimi Team, Yifan Bai, Yiping Bao, Y.~Charles, Cheng Chen, Guanduo Chen, Haiting
  Chen, Huarong Chen, Jiahao Chen, Ningxin Chen, Ruijue Chen, Yanru Chen,
  Yuankun Chen, Yutian Chen, Zhuofu Chen, Jialei Cui, Hao Ding, Mengnan Dong,
  Angang Du, and 181 others. 2025{\natexlab{a}}.
\newblock \href {https://arxiv.org/abs/2507.20534} {Kimi k2: Open agentic
  intelligence}.
\newblock \emph{Preprint}, arXiv:2507.20534.

\bibitem[{Team et~al.(2025{\natexlab{b}})Team, Li, Zhang, Zhang, Huang, Li,
  Chen, Yin, Wu, Zhou, Li, Su, Ou, Zhang, Xie, Ye, Yin, Yu, Wang, Wu, Chen,
  Zhao, Zhang, Tao, Zhang, Qiao, Wang, Yu, Fu, Shen, Yang, Lin, Zhang, Zeng,
  Yang, Yin, Song, Yan, Liao, Xia, Xiao, Min, Ding, Fang, Chen, Huang, Wang,
  Cai, Shen, Wang, Guan, Geng, Shi, Wu, Chen, Li, and
  Jiang}]{tongyideepresearch2025}
Tongyi~DeepResearch Team, Baixuan Li, Bo~Zhang, Dingchu Zhang, Fei Huang,
  Guangyu Li, Guoxin Chen, Huifeng Yin, Jialong Wu, Jingren Zhou, Kuan Li,
  Liangcai Su, Litu Ou, Liwen Zhang, Pengjun Xie, Rui Ye, Wenbiao Yin, Xinmiao
  Yu, Xinyu Wang, and 38 others. 2025{\natexlab{b}}.
\newblock \href {https://arxiv.org/abs/2510.24701} {Tongyi deepresearch
  technical report}.
\newblock \emph{Preprint}, arXiv:2510.24701.

\bibitem[{Thirukovalluru et~al.(2024)Thirukovalluru, Huang, and
  Dhingra}]{thirukovalluru2024atomicsc}
Raghuveer Thirukovalluru, Yukun Huang, and Bhuwan Dhingra. 2024.
\newblock Atomic self-consistency for better long form generations.
\newblock In \emph{Proceedings of the 2024 Conference on Empirical Methods in
  Natural Language Processing}, pages 12681--12694.

\bibitem[{Tian et~al.(2023)Tian, Mitchell, Zhou, Sharma, Rafailov, Yao, Finn,
  and Manning}]{tian2023justaskcalibrated}
Katherine Tian, Eric Mitchell, Allan Zhou, Archit Sharma, Rafael Rafailov,
  Huaxiu Yao, Chelsea Finn, and Christopher~D Manning. 2023.
\newblock Just ask for calibration: Strategies for eliciting calibrated
  confidence scores from language models fine-tuned with human feedback.
\newblock In \emph{Proceedings of the 2023 Conference on Empirical Methods in
  Natural Language Processing}, pages 5433--5442.

\bibitem[{Wang et~al.(2024)Wang, Prasad, Stengel-Eskin, and
  Bansal}]{wang2024softsc}
Han Wang, Archiki Prasad, Elias Stengel-Eskin, and Mohit Bansal. 2024.
\newblock Soft self-consistency improves language models agents.
\newblock In \emph{Proceedings of the 62nd Annual Meeting of the Association
  for Computational Linguistics (Volume 2: Short Papers)}, pages 287--301.

\bibitem[{Wang et~al.(2025)Wang, Wang, Athiwaratkun, Zhang, and
  Zou}]{wang2024mixtureofagents}
Junlin Wang, Jue Wang, Ben Athiwaratkun, Ce~Zhang, and James~Y Zou. 2025.
\newblock Mixture-of-agents enhances large language model capabilities.
\newblock In \emph{International Conference on Learning Representations},
  volume 2025, pages 33944--33963.

\bibitem[{Wang et~al.(2022{\natexlab{a}})Wang, Yang, Huang, Jiao, Yang, Jiang,
  Majumder, and Wei}]{wang2022e5}
Liang Wang, Nan Yang, Xiaolong Huang, Binxing Jiao, Linjun Yang, Daxin Jiang,
  Rangan Majumder, and Furu Wei. 2022{\natexlab{a}}.
\newblock Text embeddings by weakly-supervised contrastive pre-training.
\newblock \emph{arXiv preprint arXiv:2212.03533}.

\bibitem[{Wang et~al.(2022{\natexlab{b}})Wang, Wei, Schuurmans, Le, Chi,
  Narang, Chowdhery, and Zhou}]{wang2022selfconsistency}
Xuezhi Wang, Jason Wei, Dale Schuurmans, Quoc Le, Ed~Chi, Sharan Narang,
  Aakanksha Chowdhery, and Denny Zhou. 2022{\natexlab{b}}.
\newblock Self-consistency improves chain of thought reasoning in language
  models.
\newblock \emph{arXiv preprint arXiv:2203.11171}.

\bibitem[{Wei et~al.(2025)Wei, Sun, Papay, McKinney, Han, Fulford, Chung,
  Passos, Fedus, and Glaese}]{wei2025browsecomp}
Jason Wei, Zhiqing Sun, Spencer Papay, Scott McKinney, Jeffrey Han, Isa
  Fulford, Hyung~Won Chung, Alex~Tachard Passos, William Fedus, and Amelia
  Glaese. 2025.
\newblock Browsecomp: A simple yet challenging benchmark for browsing agents.
\newblock \emph{arXiv preprint arXiv:2504.12516}.

\bibitem[{Xiong et~al.(2024)Xiong, Hu, Lu, Li, Fu, He, and
  Hooi}]{xiong2023canllmexpress}
Miao Xiong, Zhiyuan Hu, Xinyang Lu, Yifei Li, Jie Fu, Junxian He, and Bryan
  Hooi. 2024.
\newblock Can llms express their uncertainty? an empirical evaluation of
  confidence elicitation in llms.
\newblock In \emph{International Conference on Learning Representations},
  volume 2024, pages 23650--23678.

\bibitem[{Xuan et~al.(2026)Xuan, Zeng, Qi, Xiao, Wang, and
  Yokoya}]{xuan2026dichotomy}
Weihao Xuan, Qingcheng Zeng, Heli Qi, Yunze Xiao, Junjue Wang, and Naoto
  Yokoya. 2026.
\newblock \href {https://arxiv.org/abs/2601.07264} {The confidence dichotomy:
  Analyzing and mitigating miscalibration in tool-use agents}.
\newblock \emph{Preprint}, arXiv:2601.07264.

\bibitem[{Yang et~al.(2018)Yang, Qi, Zhang, Bengio, Cohen, Salakhutdinov, and
  Manning}]{yang-etal-2018-hotpotqa}
Zhilin Yang, Peng Qi, Saizheng Zhang, Yoshua Bengio, William Cohen, Ruslan
  Salakhutdinov, and Christopher~D. Manning. 2018.
\newblock \href {https://doi.org/10.18653/v1/D18-1259} {{H}otpot{QA}: A dataset
  for diverse, explainable multi-hop question answering}.
\newblock In \emph{Proceedings of the 2018 Conference on Empirical Methods in
  Natural Language Processing}, pages 2369--2380, Brussels, Belgium.
  Association for Computational Linguistics.

\bibitem[{Yao et~al.(2023)Yao, Zhao, Yu, Du, Shafran, Narasimhan, and
  Cao}]{yao2022react}
Shunyu Yao, Jeffrey Zhao, Dian Yu, Nan Du, Izhak Shafran, Karthik Narasimhan,
  and Yuan Cao. 2023.
\newblock \href {https://arxiv.org/abs/2210.03629} {React: Synergizing
  reasoning and acting in language models}.
\newblock \emph{Preprint}, arXiv:2210.03629.

\bibitem[{Zhou et~al.(2024)Zhou, Pujara, Ren, Chen, Cheng, Le, Zhou, Mishra,
  Zheng et~al.}]{zhou2024selfdiscover}
Pei Zhou, Jay Pujara, Xiang Ren, Xinyun Chen, Heng-Tze Cheng, Quoc~V Le, Denny
  Zhou, Swaroop Mishra, Huaixiu~S Zheng, and 1 others. 2024.
\newblock Self-discover: Large language models self-compose reasoning
  structures.
\newblock \emph{Advances in Neural Information Processing Systems},
  37:126032--126058.

\end{thebibliography}

\appendix
\clearpage

\section{Implementation details}
\label{app:impl}

\paragraph{Infrastructure.}
Tongyi-DeepResearch rollouts are generated locally on
$8\times$ NVIDIA RTX PRO 6000 GPUs; all other models are served
via Fireworks AI.\footnote{\url{https://fireworks.ai}}
Web search is powered by the Serper
API\footnote{\url{https://serper.dev}} and page visits use
Crawl4AI.\footnote{\url{https://github.com/unclecode/crawl4ai}}

\paragraph{Token sets for RGV.}
Given a text span $X$, we form the token set $\mathcal{T}(X)$ by
(i)~lowercasing and Unicode-NFKC-normalising the string;
(ii)~stripping markdown markup, URLs, code fences, and JSON
delimiters; (iii)~whitespace-splitting; (iv)~removing a small
English stopword list of $\sim 130$ function words; and
(v)~dropping any token whose normalised length is below two
characters or whose normalised form is purely punctuation. No
lemmatisation or stemming is used. Each document $d_j$ is taken
to be the raw text returned by the tool, \emph{not} the model's quoted snippets, so
$\mathcal{T}(d)$ reflects what the corpus served and is
independent of how the model rendered it.

\paragraph{Computing the max-over-docs.}
For each rollout we cache $\mathcal{T}(P_i)$ once and iterate
over $\{d_1,\dots,d_{m_i}\}$, computing
$|\mathcal{T}(P_i)\cap\mathcal{T}(d)|/|\mathcal{T}(P_i)|$ via a
single set intersection per document. End-to-end runtime is
$O(|P_i| + \sum_j |d_j|)$ and finishes in $\sim 0.3$\,ms per
rollout on one CPU thread for a typical BrowseComp-Plus rollout.

\paragraph{Strict answer clustering.}
Predicted answer strings are normalised by lowercasing,
collapsing whitespace, stripping a leading article
(\textit{the/a/an}) and surrounding punctuation, and removing
matching outer quotes. Two rollouts join the same cluster iff
their normalised strings are byte-equal.

\paragraph{DeepConf computation.}
Per-token confidence is $C_t = -\tfrac{1}{k}\sum_{j=1}^{k}\log p_j$,
the negated mean log-prob over the top-$k$ candidates at step
$t$. We
compute $C_t$ over all of the rollout's generated tokens
(reasoning and answer). The headline DeepConf score is the
\emph{Lowest-Group} reduction with window size $W{=}1024$: slide
a window of $1024$ tokens with stride $1$ across the generated
sequence, compute the mean $C_t$ inside each window, and take the
minimum. Other
reductions (Bottom-$10\%$, Tail) and window sizes
$W\!\in\!\{1024, 2048, 4096\}$ are reported in
Appendix~\ref{app:dc-variants}.

\paragraph{Force-final and recovery.}
Inside the rollout loop, before each LLM call, we check whether
the prompt token count exceeds $85\%$ of the model's context
window. If so, the last assistant content is replaced with a
truncation marker and a single follow-up call is issued with
tools disabled and a markdown final-answer instruction (verbatim
in Appendix~\ref{app:examples}). The same force-final is
triggered when the agent exhausts \texttt{max\_iterations}
($200$), the cumulative output budget
(\texttt{max\_total\_output\_tokens}, $30$K). No post-hoc compressed recovery
is performed: if the single force-final call itself fails, the
rollout is recorded with status \texttt{api\_error} and an empty
\texttt{final\_text}, and is treated as \texttt{parse\_error} by
the judge.

\paragraph{Per-model sampling parameters.}
The same protocol is used across all models. Parameters not listed Table~\ref{tab:model-params} take standard
defaults. The judge is Qwen3-32B at temperature $0.0$.

\begin{table*}[t]
\centering
\small
\setlength{\tabcolsep}{6pt}
\begin{tabular}{l c c c c c}
\toprule
Parameter & gpt-oss-120b & MiniMax-M2.7 & GLM-5.1 & Kimi-K2.5 &
   Tongyi-DeepResearch \\
\midrule
rollout temperature      & 0.85 & 0.85 & 0.85 & 0.85 & 0.85 \\
top\_p                   & 1.0  & 1.0  & 1.0  & 1.0  & 1.0  \\
presence penalty         & 0.0  & 0.0  & 0.0  & 0.0  & 1.1  \\
max output tokens / turn & 8,192 & 8,192 & 8,192 & 8,192 & 8,192 \\
max total output budget  & 30,000 & 30,000 & 30,000 & 30,000 & 30,000 \\
max tool calls           & 200  & 200  & 200  & 200  & 200  \\
context window           & 131,072 & 196,607 & 202,752 & 256,000 & 131,072 \\
\bottomrule
\end{tabular}
\caption{Per-model sampling parameters. All models use the same
system prompt (Appendix~\ref{app:examples}), confidence solicitation
protocol, and Qwen3-32B judge at temperature~$0.0$.}
\label{tab:model-params}
\end{table*}

% ======================================================================
\section{Variance estimation via 3-fold split}
\label{app:variance}

The headline Table~\ref{tab:headline} reports a single mean
accuracy per cell over the $150$-question sample. To attach
uncertainty without changing the protocol, we split each
dataset's $150$ questions into three disjoint folds of $50$
(strata by the random seed already used for sampling: folds use
the first $50$, next $50$, and last $50$ qids in their sampling
order), recompute each voting method's accuracy on each fold,
and report the fold-mean $\pm$ fold-standard-deviation in
Table~\ref{tab:headline-ci} below. Folds are fixed across
methods so that paired-fold differences are well-defined.

\begin{table*}[t]
\centering
\small
\setlength{\tabcolsep}{4pt}
\begin{tabular}{l l cccc}
\toprule
\textbf{Dataset} & \textbf{Model} &
   Simple Maj. & DeepConf & RGV & Oracle \\
\midrule
\multirow{5}{*}{BrowseComp-Plus}
  & Tongyi-DeepResearch & $62.0{\pm}1.1$ & $65.7{\pm}1.0$ & $\textbf{71.1}{\pm}1.9$ & $74.2{\pm}0.5$ \\
  & OSS-120B            & $53.3{\pm}4.7$ & $54.7{\pm}4.1$ & $\textbf{56.0}{\pm}2.8$ & $66.7{\pm}4.1$ \\
  & MiniMax-M2.7        & $75.3{\pm}6.8$ & $75.3{\pm}7.4$ & $\textbf{78.7}{\pm}8.2$ & $83.3{\pm}1.9$ \\
  & Kimi-K2.5           & $80.7{\pm}5.0$ & $80.7{\pm}4.1$ & $\textbf{81.3}{\pm}5.7$ & $88.7{\pm}0.9$ \\
  & GLM-5.1             & $80.7{\pm}4.5$ & $80.7{\pm}5.2$ & $\textbf{82.0}{\pm}4.8$ & $90.0{\pm}1.6$ \\
\cmidrule(lr){1-6}
\multirow{5}{*}{GAIA}
  & Tongyi-DeepResearch & $79.6{\pm}5.8$ & $79.6{\pm}5.8$ & $\textbf{80.6}{\pm}3.9$ & $92.2{\pm}4.2$ \\
  & OSS-120B            & $69.9{\pm}3.1$ & $68.9{\pm}4.5$ & $\textbf{71.8}{\pm}3.8$ & $83.5{\pm}2.9$ \\
  & MiniMax-M2.7        & $87.4{\pm}3.6$ & $88.3{\pm}5.0$ & $\textbf{93.2}{\pm}2.7$ & $97.1{\pm}2.4$ \\
  & Kimi-K2.5           & $78.6{\pm}3.7$ & $79.6{\pm}4.8$ & $\textbf{81.6}{\pm}2.1$ & $92.2{\pm}2.8$ \\
  & GLM-5.1             & $87.4{\pm}0.2$ & $87.4{\pm}2.4$ & $\textbf{89.3}{\pm}2.2$ & $92.2{\pm}0.0$ \\
\cmidrule(lr){1-6}
\multirow{5}{*}{BrowseComp}
  & Tongyi-DeepResearch & $52.0{\pm}2.8$ & $52.7{\pm}3.8$ & $\textbf{54.0}{\pm}2.5$ & $68.7{\pm}3.4$ \\
  & OSS-120B            & $30.0{\pm}0.9$ & $29.3{\pm}0.9$ & $\textbf{34.0}{\pm}1.6$ & $41.3{\pm}3.4$ \\
  & MiniMax-M2.7        & $40.7{\pm}5.2$ & $46.0{\pm}4.3$ & $\textbf{50.7}{\pm}6.2$ & $59.3{\pm}3.8$ \\
  & Kimi-K2.5           & $52.8{\pm}10.4$ & $56.6{\pm}8.1$ & $\textbf{58.5}{\pm}7.8$ & $73.6{\pm}13.3$ \\
  & GLM-5.1             & $72.6{\pm}6.0$ & $73.7{\pm}5.3$ & $\textbf{78.9}{\pm}5.3$ & $92.6{\pm}0.9$ \\
\cmidrule(lr){1-6}
\multirow{5}{*}{FRAMES}
  & Tongyi-DeepResearch & $87.3{\pm}3.1$ & $88.0{\pm}2.8$ & $\textbf{88.0}{\pm}3.5$ & $94.7{\pm}2.1$ \\
  & OSS-120B            & $82.0{\pm}4.7$ & $82.7{\pm}5.0$ & $\textbf{85.3}{\pm}5.7$ & $91.3{\pm}3.4$ \\
  & MiniMax-M2.7        & $88.7{\pm}1.9$ & $89.3{\pm}2.8$ & $\textbf{89.3}{\pm}2.5$ & $94.0{\pm}1.6$ \\
  & Kimi-K2.5           & $88.0{\pm}2.8$ & $88.7{\pm}3.4$ & $\textbf{89.3}{\pm}4.7$ & $97.3{\pm}1.3$ \\
  & GLM-5.1             & $90.0{\pm}2.8$ & $90.0{\pm}2.8$ & $\textbf{90.7}{\pm}1.9$ & $98.0{\pm}0.9$ \\
\bottomrule
\end{tabular}
\caption{Headline accuracies (\%) with fold-mean$\pm$std under a
3-fold split of each benchmark's question pool.}
\label{tab:headline-ci}
\end{table*}

\noindent
This is a coarse estimator: with three folds the standard
deviation is itself noisy. We use it as a stability indicator
(does the margin survive a different 50-question slice?) rather
than for formal hypothesis tests. The point estimates in
Table~\ref{tab:headline} are computed on all 150 questions.

% ======================================================================
\section{Verbalised confidence as a voting weight}
\label{app:verb}

The main text compares RGV against logprob-based confidence
(DeepConf), which is the dominant paradigm in the voting literature
\citep{fu2025deepconf,wang2022selfconsistency,snell2024scalecompute}.
A natural question is how \emph{verbalised} confidence---asking the
model to self-report a numerical confidence
score---performs as an alternative voting weight
\citep{taubenfeld2025cisc,xuan2026dichotomy,ou2025browseconf}.
Verbalised confidence is arguably closer to an ``outside-context''
signal than raw logprobs, since it does not directly read
per-token probabilities from the contaminated generation;
however, the self-assessment is still produced by the same model
reasoning under the same context, so it remains susceptible to the
overconfidence patterns documented by
\citet{xuan2026dichotomy}. We include it here for completeness
as a third signal class, distinct from both the logit-based
(DeepConf) and the retrieval-based (RGV) families.

\paragraph{Verbalised-confidence elicitation.}
After each rollout's final answer is produced, the agent is
issued one deterministic follow-up call at temperature $0$, with
tools disabled (\texttt{tool\_choice="none"}) and
\texttt{max\_tokens=1024}. The prompt asks for a single integer between
$0$ and $100$ representing the verbalised confidence that the
just-given answer is correct (verbatim text in
Appendix~\ref{app:examples}). The reply is parsed for the last
labelled or bare integer in $[0,100]$ and stored as a
per-rollout score; this score is used only in
Appendix~\ref{app:verb}.

The protocol elicits a verbalised confidence $v_i\!\in\![0,100]$
on every rollout (\S\ref{sec:setup}, Appendix~\ref{app:examples}).
We report it separately from the main-text comparison to keep
the signal-class contrast clean (logit-based vs.\
retrieval-based); this section evaluates it as a voting weight
under the same weighted majority rule as the other methods.

\begin{table*}[t]
\centering
\small
\setlength{\tabcolsep}{4pt}
\begin{tabular}{l l cccc cc}
\toprule
\textbf{Dataset} & \textbf{Model} &
   SM & DeepConf & \textsc{Verb} & RGV & DeepConf+RGV & Verb+RGV \\
\midrule
\multirow{5}{*}{BrowseComp-Plus}
  & Tongyi-DeepResearch & 62.0 & 65.7 & 68.0 & 71.1 & 70.7 & \textbf{71.3} \\
  & OSS-120B            & 53.3 & 54.7 & 56.7 & 56.0 & 56.0 & \textbf{58.0} \\
  & MiniMax-M2.7        & 75.3 & 75.3 & 74.0 & \textbf{78.7} & 78.0 & 78.0 \\
  & Kimi-K2.5           & 80.7 & 80.7 & 80.0 & 81.3 & 81.3 & \textbf{82.0} \\
  & GLM-5.1             & 80.7 & 80.7 & 82.0 & 82.0 & 82.0 & \textbf{82.7} \\
\cmidrule(lr){1-8}
\multirow{5}{*}{GAIA}
  & Tongyi-DeepResearch & 79.6 & 79.6 & 73.8 & \textbf{80.6} & \textbf{80.6} & 79.6 \\
  & OSS-120B            & 69.9 & 68.9 & 71.8 & 71.8 & 71.8 & \textbf{72.8} \\
  & MiniMax-M2.7        & 87.4 & 88.3 & 91.3 & \textbf{93.2} & 92.2 & \textbf{93.2} \\
  & Kimi-K2.5           & 78.6 & 79.6 & 82.5 & 81.6 & 82.5 & \textbf{85.4} \\
  & GLM-5.1             & 87.4 & 87.4 & 86.4 & \textbf{89.3} & \textbf{89.3} & 88.3 \\
\cmidrule(lr){1-8}
\multirow{5}{*}{BrowseComp}
  & Tongyi-DeepResearch & 52.0 & 52.7 & 53.3 & \textbf{54.0} & 53.3 & \textbf{54.0} \\
  & OSS-120B            & 30.0 & 29.3 & \textbf{35.3} & 34.0 & 34.0 & 34.7 \\
  & MiniMax-M2.7        & 40.7 & 46.0 & 48.0 & 50.7 & \textbf{52.7} & 52.0 \\
  & Kimi-K2.5           & 52.8 & 56.6 & 60.4 & 58.5 & 60.4 & \textbf{62.3} \\
  & GLM-5.1             & 72.6 & 73.7 & \textbf{78.9} & \textbf{78.9} & 74.7 & \textbf{78.9} \\
\cmidrule(lr){1-8}
\multirow{5}{*}{FRAMES}
  & Tongyi-DeepResearch & 87.3 & 88.0 & 87.3 & \textbf{88.0} & \textbf{88.0} & 87.3 \\
  & OSS-120B            & 82.0 & 82.7 & 82.7 & \textbf{85.3} & 84.7 & 84.0 \\
  & MiniMax-M2.7        & 88.7 & 89.3 & 88.0 & \textbf{89.3} & \textbf{89.3} & 89.3 \\
  & Kimi-K2.5           & 88.0 & 88.7 & 84.0 & \textbf{89.3} & 88.7 & 88.0 \\
  & GLM-5.1             & 90.0 & 90.0 & 90.0 & \textbf{90.7} & 90.0 & \textbf{90.7} \\
\bottomrule
\end{tabular}
\caption{Voting accuracies (\%) including the verbalised-confidence
weight (\textsc{Verb}).}
\label{tab:verb}
\end{table*}

\noindent
Verbalised confidence occupies an intermediate position: it does
not directly read per-token logprobs, but the self-assessment is
still produced under the same context that contains the retrieved
documents. As \citet{xuan2026dichotomy} and
\citet{ou2025browseconf} also observe, this makes it partially
susceptible to the overconfidence patterns of
\S\ref{sec:motivation}, though to a lesser degree than
logprob-based signals.

\paragraph{Combining signals.}
The DC+RGV and Verb+RGV columns in Table~\ref{tab:verb} report a
simple combination: within each question, we $z$-normalise each
signal (zero mean, unit variance) and form the voting weight
$w_i = z(\text{RGV}_i) + \alpha\, z(\text{other}_i)$, where
$\alpha$ is selected from $\{0.05, 0.1, 0.15, 0.2, 0.3, 0.5, 1.0\}$
per cell to maximise voting accuracy. Verb+RGV improves over RGV alone in $9$ of $20$ cells (up to
$+3.8\%$ on GAIA Kimi-K2.5 and BrowseComp Kimi-K2.5), suggesting
that verbalised confidence captures signal orthogonal to lexical
grounding. DC+RGV improves in $3$ of $20$ cells, indicating that
logprob-based confidence is largely subsumed once grounding is
accounted for. The Verb+RGV combination is a promising direction
for settings where a follow-up confidence call is affordable.

% ======================================================================
\section{Cross-dataset evidence for copy-inflation}
\label{app:cross-dataset}

The mechanism analysis in \S\ref{sec:motivation} uses
BrowseComp-Plus with Tongyi-DeepResearch as the deep-dive
configuration. To verify that the copy-inflation pattern is not
specific to that setting, we replicate the key diagnostic figures
across all twenty benchmark$\times$model cells.

\begin{figure*}[t]
  \centering
  \includegraphics[width=\textwidth]{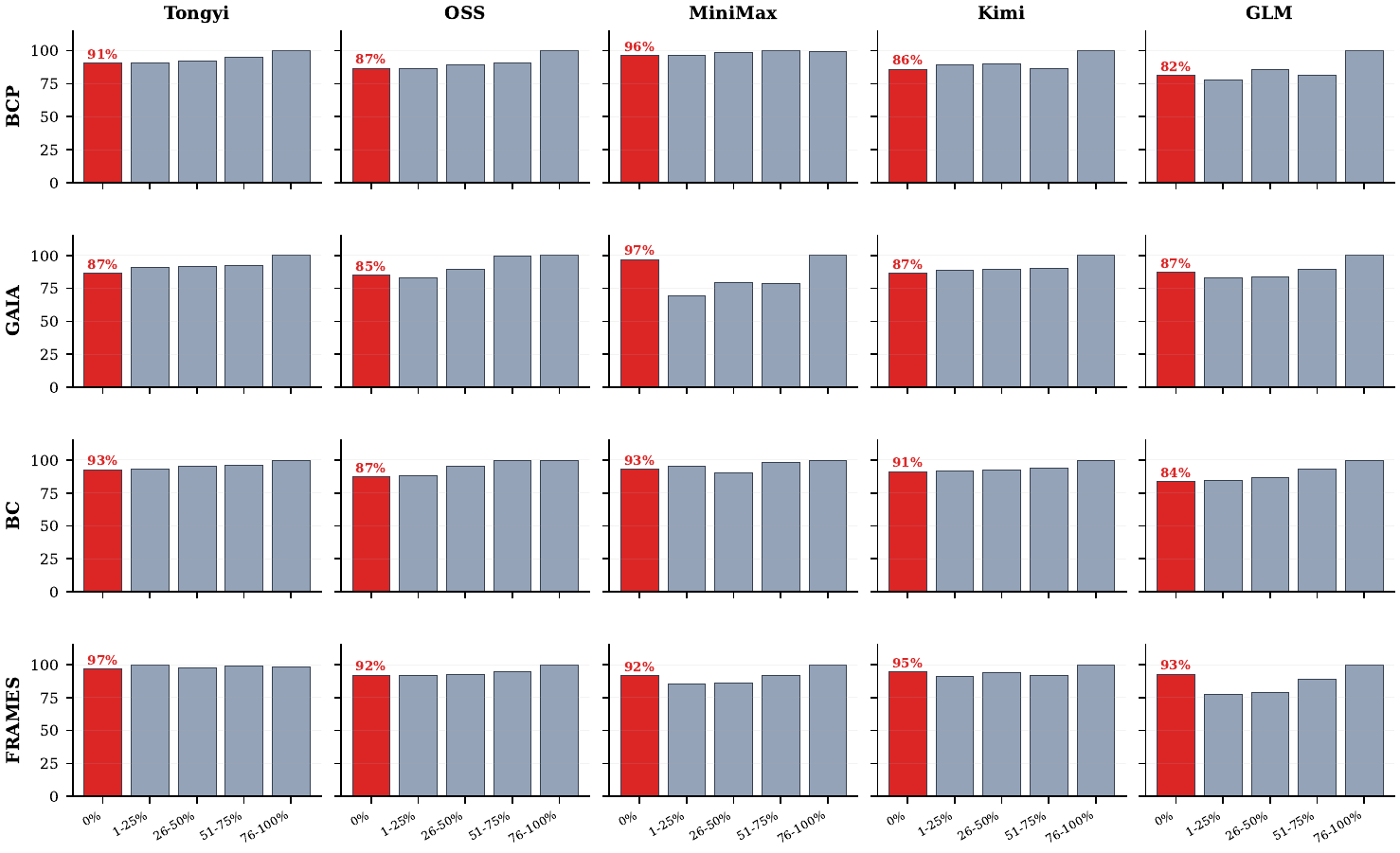}
  \caption{\textbf{DeepConf overconfidence across all 20
    benchmark$\times$model cells} (cf.\ Figure~\ref{fig:motivation}a).
    Each panel stratifies questions by oracle difficulty and reports
    the mean DeepConf score as a percentage of its peak. On every
    configuration, impossible questions ($0\%$ correct, red) receive
    $82$--$97\%$ of the peak score. Rows: datasets; columns: models.}
  \label{fig:cross-overconf}
\end{figure*}

Figure~\ref{fig:motivation}(a) showed that DeepConf gives impossible
questions $87\%$ of its peak score on BrowseComp-Plus with Tongyi.
Figure~\ref{fig:cross-overconf} extends this across all twenty
configurations. On every combination, impossible questions receive
$82$--$97\%$ of the peak, confirming that copy-inflation is a
general property of logprob-based confidence in multi-turn search
agents, not an artefact of one benchmark or model.

\paragraph{Severity across model families.}
Figure~\ref{fig:cross-overconf} shows the behavioural signature; the
mechanism itself can be measured directly. Table~\ref{tab:families}
reports, per model family, the median fraction of generated tokens
that are copies and the copy-vs-non-copy logprob gap, adding a
DeepSeek cell (deepseek-v4-flash on GAIA, $103$ questions $\times$ $8$)
to the five families of Table~\ref{tab:headline}.

\begin{table}[h]
\centering\small
\setlength{\tabcolsep}{4pt}
\begin{tabular}{l l c c}
\toprule
\textbf{Family} & \textbf{Cell} & copy \% & logprob gap \\
\midrule
GPT-OSS  & FRAMES          & 82.1 & $+0.15$ \\
GPT-OSS  & BrowseComp-Plus & 91.9 & $+0.07$ \\
Tongyi   & FRAMES          & 85.2 & $+0.69$ \\
Tongyi   & BrowseComp-Plus & 91.9 & $+0.11$ \\
GLM      & FRAMES          & 87.8 & $+0.16$ \\
Kimi     & FRAMES          & 87.7 & $+0.20$ \\
MiniMax  & FRAMES          & 84.8 & $+0.40$ \\
DeepSeek & GAIA            & 89.7 & $+0.04$ \\
\bottomrule
\end{tabular}
\caption{\textbf{Copy-inflation severity by model family.} The gap is
positive in every cell; copying accounts for $78$--$92\%$ of generated
tokens across all six families (the $78\%$ lower end is the wiki-18
cell of Appendix~\ref{app:weak-retrieval}).}
\label{tab:families}
\end{table}

\noindent
The mechanism replicates in every family we tested, including a
reasoning-tuned DeepSeek model. Severity tracks retrieval-bag size and
tool style more closely than training objective: the DeepSeek cell has
the smallest gap ($+0.04$ nats) yet still copies $89.7\%$ of its
tokens, and the within-question flattening of
Appendix~\ref{app:interventions} holds there too. Disentangling
training objective from tool style would need controlled pairs, which
we leave to future work.

% ======================================================================
\section{Robustness to alternative lexical-overlap variants}
\label{app:lex-metrics}

The headline method (\S\ref{sec:method}) uses prose-recall under
the strict clustering rule. To check that the gain comes from
the signal source rather than the specific overlap function, we
swap the per-rollout score $w_i^{\text{RGV}}$ for other members
of the set-overlap family (and a few neighbours), keeping
everything else fixed (Table~\ref{tab:lex-variants}).

\begin{table*}[t]
\centering
\small
\setlength{\tabcolsep}{4pt}
\begin{tabular}{l l cccccc}
\toprule
\textbf{Dataset} & \textbf{Model} &
   prose-recall & Jaccard & ROUGE-2 & ROUGE-L & BM25 & TF-IDF \\
\midrule
\multirow{5}{*}{BrowseComp-Plus}
  & Tongyi-DeepResearch & 71.1 & 71.1 & 70.2 & \textbf{71.3} & \textbf{71.7} & 69.0 \\
  & OSS-120B            & 56.0 & 56.0 & \textbf{56.7} & 55.3 & 55.3 & 52.7 \\
  & MiniMax-M2.7        & \textbf{78.7} & 77.3 & 74.0 & 76.0 & 75.3 & 75.3 \\
  & Kimi-K2.5           & \textbf{81.3} & 80.0 & 79.3 & 79.3 & 77.3 & 78.7 \\
  & GLM-5.1             & \textbf{82.0} & 79.3 & 80.0 & 80.7 & 79.3 & 78.7 \\
\cmidrule(lr){1-8}
\multirow{5}{*}{GAIA}
  & Tongyi-DeepResearch & \textbf{80.6} & 70.9 & 72.8 & 71.8 & 71.8 & 73.8 \\
  & OSS-120B            & \textbf{71.8} & 69.9 & 68.9 & 68.0 & 68.0 & 67.0 \\
  & MiniMax-M2.7        & \textbf{93.2} & 91.3 & 91.3 & 91.3 & 91.3 & 89.3 \\
  & Kimi-K2.5           & \textbf{81.6} & 78.6 & 77.7 & 76.7 & 79.6 & 78.6 \\
  & GLM-5.1             & \textbf{89.3} & 88.3 & 88.3 & 87.4 & 88.3 & 87.4 \\
\cmidrule(lr){1-8}
\multirow{5}{*}{BrowseComp}
  & Tongyi-DeepResearch & \textbf{54.0} & 53.3 & \textbf{54.0} & 52.7 & 50.7 & 52.0 \\
  & OSS-120B            & \textbf{34.0} & 33.3 & 32.7 & 32.7 & 31.3 & 30.7 \\
  & MiniMax-M2.7        & \textbf{50.7} & 48.0 & 45.3 & 45.3 & 46.0 & 46.0 \\
  & Kimi-K2.5           & \textbf{58.5} & 52.8 & \textbf{58.5} & 52.8 & 50.9 & 52.8 \\
  & GLM-5.1             & 78.9 & \textbf{82.1} & 77.9 & 76.8 & 69.5 & 81.1 \\
\cmidrule(lr){1-8}
\multirow{5}{*}{FRAMES}
  & Tongyi-DeepResearch & \textbf{88.0} & \textbf{88.0} & 86.7 & 87.3 & 86.0 & 86.7 \\
  & OSS-120B            & \textbf{85.3} & 79.3 & 82.0 & 82.0 & 80.7 & 80.0 \\
  & MiniMax-M2.7        & \textbf{89.3} & \textbf{89.3} & 88.0 & 88.7 & 88.0 & 87.3 \\
  & Kimi-K2.5           & 88.7 & \textbf{89.3} & 87.3 & 88.0 & 86.7 & 87.3 \\
  & GLM-5.1             & \textbf{90.7} & 90.0 & 89.3 & 90.0 & 89.3 & 88.7 \\
\bottomrule
\end{tabular}
\caption{Voting accuracy (\%) across lexical-overlap variants on
the main cells. All variants use the same max-over-docs reduction;
only $w_i^{\text{RGV}}$ changes. The highest value in each row
is bolded.}
\label{tab:lex-variants}
\end{table*}

\noindent
\textit{Definitions.} Jaccard:
$|\mathcal{T}(P)\cap\mathcal{T}(d)|/|\mathcal{T}(P)\cup\mathcal{T}(d)|$
(symmetric variant of prose-recall). ROUGE-2: bigram $F_1$ between
prose and doc. ROUGE-L: unigram $F_1$ (approximating LCS-based
ROUGE-L). BM25: scored as the maximum BM25 score of $P$
against each $d$ under the per-rollout retrieval set as the
corpus. TF-IDF: cosine of TF-IDF vectors (same corpus).

\paragraph{A semantic scorer in place of a lexical one.}
Every variant above is lexical, which leaves open whether the gain
depends on surface overlap. We therefore replace the score with a
frozen dense encoder, \texttt{intfloat/e5-base-v2}
\citep{wang2022e5} ($109$M parameters, no fine-tuning), and set
$w_i = \max_{d \in \mathcal{D}_i} \cos\!\big(E(P_i), E(d)\big)$,
keeping the max-over-docs reduction and everything else fixed.

\begin{center}\footnotesize
\setlength{\tabcolsep}{4pt}
\begin{tabular}{lccc}
\toprule
Method & \begin{tabular}{@{}c@{}}FRAMES\\OSS\end{tabular}
       & \begin{tabular}{@{}c@{}}BM25\\OSS\end{tabular}
       & \begin{tabular}{@{}c@{}}BM25\\Tongyi\end{tabular} \\
\midrule
Simple majority   & 82.0 & 38.7 & 58.7 \\
DeepConf          & 82.7 & 40.7 & 59.3 \\
Embedding cosine  & 83.3 & 42.0 & 60.0 \\
RGV               & \textbf{85.3} & \textbf{44.0} & \textbf{61.3} \\
\bottomrule
\end{tabular}
\end{center}

\noindent
The two BM25 columns are the degraded-retrieval cells of
Appendix~\ref{app:weak-retrieval}. On all three, the embedding voter
beats both simple majority and DeepConf but does not reach the lexical
score, at roughly five orders of magnitude more compute (encoder
minutes versus ${\sim}0.3$\,ms per rollout). Two known properties of
dense encoders are consistent with the ordering: they degrade on
exactly the rare entities that carry our signal
(Figure~\ref{fig:why}(\subref{fig:why-a}))
\citep{sciavolino-etal-2021-simple}, and fixed-dimension embeddings
lose precision on long documents \citep{luan-etal-2021-sparse}; a
similar pattern is reported for hallucination detection, where an
$n$-gram check outperforms grey-box log-probability baselines in most
setups \citep{manakul-etal-2023-selfcheckgpt}.

The point is not that lexical overlap is optimal. It is that two
scorers with nothing in common except \emph{where they read from}
land on the same side of the baselines, which is what the
copy-inflation account predicts: the signal source carries the gain,
not the metric.

% ======================================================================
\section{DeepConf variant grid}
\label{app:dc-variants}

\citet{fu2025deepconf} propose three reductions of sliding-window
group confidences: \emph{Lowest-Group} (sliding-window minimum,
the extreme case of Bottom-$q\%$ as $q\!\to\!0$),
\emph{Bottom-$10\%$}, and \emph{Tail}. We compute the per-token
confidence as they do,
$C_t = -\tfrac{1}{k}\sum_{j=1}^{k}\log p_j$, and sweep all three
reductions across three window sizes
$W\!\in\!\{1024, 2048, 4096\}$ on the rollout's generated
tokens; per-rollout weights are then plugged into the weighted
majority vote of \S\ref{sec:background}
(Table~\ref{tab:dc-variants}).

\begin{table*}[t]
\centering
\small
\setlength{\tabcolsep}{3.5pt}
\begin{tabular}{l l ccc ccc ccc}
\toprule
& & \multicolumn{3}{c}{Lowest-Group ($\min$)} &
    \multicolumn{3}{c}{Bottom-10\%} &
    \multicolumn{3}{c}{Tail} \\
\cmidrule(lr){3-5}\cmidrule(lr){6-8}\cmidrule(lr){9-11}
\textbf{Dataset} & \textbf{Model} &
   1024 & 2048 & 4096 &
   1024 & 2048 & 4096 &
   1024 & 2048 & 4096 \\
\midrule
\multirow{5}{*}{BrowseComp-Plus}
  & Tongyi-DeepResearch & \textbf{65.7} & 65.4 & 64.8 & \textbf{65.7} & 65.5 & 64.9 & 61.8 & 61.9 & 61.4 \\
  & OSS-120B            & \textbf{54.7} & 54.0 & 53.3 & \textbf{54.7} & 54.0 & 53.3 & 53.3 & 52.7 & 52.0 \\
  & MiniMax-M2.7        & \textbf{75.3} & 74.7 & 74.0 & \textbf{75.3} & 74.7 & 73.3 & 73.3 & 72.7 & 72.0 \\
  & Kimi-K2.5           & \textbf{80.7} & 80.0 & 79.3 & \textbf{80.7} & 80.0 & 79.3 & 78.0 & 78.0 & 77.3 \\
  & GLM-5.1             & \textbf{80.7} & 80.0 & 79.3 & \textbf{80.7} & 80.0 & 79.3 & 78.7 & 78.0 & 77.3 \\
\cmidrule(lr){1-11}
\multirow{5}{*}{GAIA}
  & Tongyi-DeepResearch & \textbf{79.6} & 78.6 & 77.7 & \textbf{79.6} & 78.6 & 77.7 & 76.7 & 76.7 & 75.7 \\
  & OSS-120B            & \textbf{68.9} & 67.0 & 65.0 & \textbf{68.9} & 67.0 & 65.0 & 64.1 & 64.1 & 63.1 \\
  & MiniMax-M2.7        & \textbf{88.3} & 87.4 & 86.4 & \textbf{88.3} & 87.4 & 86.4 & 85.4 & 84.5 & 83.5 \\
  & Kimi-K2.5           & \textbf{79.6} & 78.6 & 77.7 & \textbf{79.6} & 78.6 & 77.7 & 76.7 & 75.7 & 74.8 \\
  & GLM-5.1             & \textbf{87.4} & 86.4 & 85.4 & \textbf{87.4} & 86.4 & 85.4 & 85.4 & 84.5 & 83.5 \\
\cmidrule(lr){1-11}
\multirow{5}{*}{BrowseComp}
  & Tongyi-DeepResearch & \textbf{52.7} & 51.3 & 50.0 & \textbf{52.7} & 51.3 & 50.0 & 49.3 & 48.7 & 47.3 \\
  & OSS-120B            & \textbf{29.3} & 28.7 & 28.0 & \textbf{29.3} & 28.7 & 28.0 & 27.3 & 27.3 & 26.7 \\
  & MiniMax-M2.7        & \textbf{46.0} & 44.7 & 43.3 & \textbf{46.0} & 44.7 & 43.3 & 42.0 & 41.3 & 40.7 \\
  & Kimi-K2.5           & \textbf{56.6} & 54.7 & 52.8 & \textbf{56.6} & 54.7 & 52.8 & 52.8 & 50.9 & 49.1 \\
  & GLM-5.1             & \textbf{73.7} & 72.6 & 71.6 & \textbf{73.7} & 72.6 & 71.6 & 69.5 & 68.4 & 67.4 \\
\cmidrule(lr){1-11}
\multirow{5}{*}{FRAMES}
  & Tongyi-DeepResearch & \textbf{88.0} & 87.3 & 86.7 & \textbf{88.0} & 87.3 & 86.7 & 86.0 & 85.3 & 84.7 \\
  & OSS-120B            & \textbf{82.7} & 82.0 & 81.3 & \textbf{82.7} & 82.0 & 81.3 & 80.0 & 80.0 & 79.3 \\
  & MiniMax-M2.7        & \textbf{89.3} & 88.7 & 88.0 & \textbf{89.3} & 88.7 & 88.0 & 86.7 & 86.0 & 85.3 \\
  & Kimi-K2.5           & \textbf{88.7} & 88.0 & 87.3 & \textbf{88.7} & 88.0 & 87.3 & 86.0 & 85.3 & 84.7 \\
  & GLM-5.1             & \textbf{90.0} & 89.3 & 88.7 & \textbf{90.0} & 89.3 & 88.7 & 88.0 & 87.3 & 86.7 \\
\bottomrule
\end{tabular}
\caption{DeepConf voting accuracy (\%) over the 3~reductions
$\times$ 3~window sizes grid (generated tokens). The
Lowest-Group $\times$ $W{=}1024$ cell matches the DC
headline in Table~\ref{tab:headline}. The highest value
in each row is bolded.}
\label{tab:dc-variants}
\end{table*}

% ======================================================================
\section{Robustness of the copy-inflation gap}
\label{app:copy}

The headline number in \S\ref{sec:motivation}
(Fig.~\ref{fig:copy}) is the per-token mean logprob gap between
copy and non-copy content tokens: $+0.50$ nats overall,
$+0.61$ on judge-correct rollouts, $+0.38$ on judge-wrong ones.
To check the gap is not an artefact of any single token-subset
choice we re-measure under four alternative rules. All
measurements use the same pool of $n{=}307$ rollouts drawn from
the BrowseComp-Plus $\times$ Tongyi-DeepResearch runs ($830$
questions, $8$ rollouts each). The pool is constructed by
uniformly sampling $50$ questions per rollout seed and keeping
those whose token-level statistics and retrieved-doc texts are
both available ($307/400$); the random seed is fixed for
reproducibility.

\paragraph{Common procedure.}
For each rollout we form two artefacts. (i)~The
\emph{retrieved-doc text} $D$: a lowercased, whitespace-collapsed
concatenation of everything the search and visit tools returned
to the agent. (ii)~The \emph{generated-token stream}: every
per-token logprob the agent emitted inside its chain-of-thought
reasoning span. Each generated token is normalised by stripping
subword-piece prefixes, lower-casing, and keeping only
alphanumeric characters; tokens that normalise to fewer than two
characters are discarded. A token is a \emph{copy token} if its
normalised surface is a substring of $D$, else \emph{non-copy}.
Substring matching makes the test robust to BPE splits of rare
entities.

For a token-subset rule with predicate $\phi(\cdot)$ and weight
$\omega(\cdot)$ we pool every kept token across rollouts and
report
\begin{equation*}
\begin{aligned}
\Delta \;=\;
&\frac{\sum_{t\in\mathrm{copy},\,\phi(t)} \omega(t)\log p(t)}
      {\sum_{t\in\mathrm{copy},\,\phi(t)} \omega(t)} \\
\;-\;
&\frac{\sum_{t\in\overline{\mathrm{copy}}\,,\phi(t)} \omega(t)\log p(t)}
      {\sum_{t\in\overline{\mathrm{copy}}\,,\phi(t)} \omega(t)},
\end{aligned}
\end{equation*}
the same token-level aggregation as the main-text headline of
$+0.50$ nats.

\paragraph{Variants.}
\begin{itemize}
  \item \textbf{All alnum tokens}: $\phi\equiv\mathrm{True}$,
    $\omega\equiv 1$. Baseline; reproduces the main-text headline.
  \item \textbf{Stopword-removed}: $\phi(t)=[w_t\notin\mathcal{S}]$
    with $\mathcal{S}$ a $\sim 120$-word English stopword list.
  \item \textbf{IDF-weighted}: $\phi$ as in stopword-removed,
    $\omega(t)=\log(N/\mathrm{df}(w_t))$ with sample IDF over
    $N{=}307$ documents-as-rollouts.
  \item \textbf{Digits or capitalised-leading}: digit tokens of
    length $\geq 2$ or BPE-stripped tokens starting uppercase,
    $\omega\equiv 1$.
  \item \textbf{Length} $\geq 8$: $\phi(t)=[|w_t|\geq 8]$,
    $\omega\equiv 1$.
\end{itemize}

\begin{table}[h]
\centering
\small
\setlength{\tabcolsep}{4pt}
\begin{tabular}{l r r r}
\toprule
Token subset & $|\mathrm{copy}|$ & $|\overline{\mathrm{copy}}|$ & gap \\
\midrule
all alnum (baseline)      & 936\,332 & 84\,584 & $+0.51$ \\
stopword-removed          & 558\,491 & 74\,714 & $+0.54$ \\
IDF-weighted              & 558\,491 & 74\,714 & $+0.69$ \\
digits / cap-leading      & 232\,855 & 21\,690 & $+0.44$ \\
length $\geq 8$           &  92\,233 & 25\,935 & $+0.62$ \\
\bottomrule
\end{tabular}
\caption{Copy-vs-non-copy logprob gap under five token-subset
rules. All gaps positive at $p\ll 0.001$ (paired bootstrap
over rollouts).}
\label{tab:copy-robust}
\end{table}

\noindent
All five rules produce a significantly positive gap, confirming
that the copy-inflation gap is not a side-effect of any
particular token-subset choice. The gap is largest under
IDF-weighting, consistent with the visual story that rare,
entity-like tokens (the kind retrieval brings into context) are
exactly the ones the agent copies.

% ======================================================================
\section{Degraded retrieval: weaker retriever and corpus mismatch}
\label{app:weak-retrieval}

RGV reads its signal from the retrieval log, so its behaviour under a
weaker retrieval stack is a boundary worth measuring rather than
assuming. We report four additional cells, each $150$ questions
$\times$ $8$ rollouts under the protocol of \S\ref{sec:setup} with the
same Qwen3-32B judge, varying only the retrieval component.

\paragraph{Two degradation modes.}
In the first, we replace the dense retriever
(Qwen3-Embedding-8B) on BrowseComp-Plus with BM25
\citep{robertson2009bm25}, holding the corpus and every other part of
the agent fixed. In the second, we adopt the Search-R1 retrieval stack
\citep{jin2025searchr1}: the wiki-18 corpus ($21$M passages) with the
E5 retriever \citep{wang2022e5}, top-$5$ passages per call. We run
that stack both on HotpotQA \citep{yang-etal-2018-hotpotqa}, whose
questions are Wikipedia-native, and on the same FRAMES questions used
in Table~\ref{tab:headline}, where the corpus no longer covers the
questions.

\begin{table*}[t]
\centering
\small
\setlength{\tabcolsep}{4pt}
\begin{tabular}{l l ccc cc c}
\toprule
\textbf{Retrieval stack} & \textbf{Model} & Single &
   SM & DeepConf & RGV$_{\text{prose}}$ & RGV$_{\text{Jac}}$ & Oracle \\
\midrule
BM25 $\leftarrow$ dense (BCP+)   & gpt-oss-120b        & 31.4 \small{(47.3)} & 38.7 & 40.7 & 37.3 & \textbf{44.0} & 52.7 \\
BM25 $\leftarrow$ dense (BCP+)   & Tongyi-DeepResearch & 48.8 \small{(58.9)} & 58.7 & 59.3 & 60.7 & \textbf{61.3} & 79.3 \\
\cmidrule(lr){1-8}
wiki-18 + E5, HotpotQA           & gpt-oss-120b        & 72.7 & \textbf{74.7} & 75.3 & 73.3 & 72.7 & 80.7 \\
wiki-18 + E5, FRAMES questions   & gpt-oss-120b        & 59.6 \small{(78.7)} & 58.7 & 58.0 & \textbf{60.0} & 59.3 & 72.0 \\
\bottomrule
\end{tabular}
\caption{\textbf{Voting accuracy (\%) under degraded retrieval.}
Parenthesised values are the single-rollout accuracy of the
corresponding stock cell, i.e.\ how far the swap moved the agent.
We report both members of the overlap family
(Appendix~\ref{app:lex-metrics}). Length-weighted voting scores
$38.0$, $56.0$, $74.7$ and $60.7$ on the four rows respectively.}
\label{tab:weak-retrieval}
\end{table*}

\paragraph{Where degraded retrieval bites, grounding wins.}
On the two BM25 rows the swap genuinely degrades the agent (single
accuracy $47.3\!\to\!31.4$ and $58.9\!\to\!48.8$), and the grounded
vote keeps a clear margin: $+5.3/+3.3$ over simple majority and
DeepConf on gpt-oss-120b, and $+2.6/+2.0$ on Tongyi-DeepResearch.

\paragraph{Where it does not, methods converge.}
On HotpotQA the Search-R1 stack turns out to be sufficient for
Wikipedia-native questions: single accuracy is $72.7$ with only
$8$ points of headroom to the oracle, and all four aggregation rules
sit inside a $2.6$-point band. This is also the lightest-copying cell
we measured ($78.4\%$ of generated tokens, versus $91.9\%$ on
BrowseComp-Plus) and the one with the healthiest DeepConf
(rollout-level AUC $0.82$), exactly as the copy-inflation account of
\S\ref{sec:motivation} predicts: less copying, less contamination,
more usable internal confidence.

\paragraph{Under corpus mismatch, only grounding stays above single.}
Running the same stack on FRAMES questions is the harshest setting:
retrieval falls out of domain and single-rollout accuracy drops
$78.7\!\to\!59.6$. Here simple majority ($58.7$) and DeepConf
($58.0$) fall \emph{below} the single-rollout average, because
errors become correlated when every rollout retrieves from the wrong
corpus, and a vote over correlated errors amplifies rather than
cancels them. RGV ($60.0$) is the only aggregation rule that stays
above it.

Taken together, degradation is graceful and, more usefully, it is
predictable in advance: the copy fraction of a cell and its remaining
headroom to the oracle tell which regime applies before any voting
rule is chosen.

% ======================================================================
\section{Causal interventions on copy-inflation}
\label{app:interventions}

\S\ref{sec:motivation} establishes copy-inflation by correlation:
copy-heavy rollouts have flatter DeepConf scores. This appendix
reports two interventions that test the causal direction.

\subsection{Intervention 1: masking copied tokens}

If copying causes the flattening, then recomputing DeepConf on
\emph{non-copy tokens only} should restore within-question spread.
And if the internal signal were merely diluted rather than corrupted,
discrimination should return along with the spread. We recompute
DeepConf over the complement of the copy set (the substring rule of
Appendix~\ref{app:copy}) on ten cells spanning six model families and
four corpora.

\begin{table*}[t]
\centering
\small
\setlength{\tabcolsep}{5pt}
\begin{tabular}{l l c cc cc c}
\toprule
& & & \multicolumn{2}{c}{rollout AUC} &
    \multicolumn{2}{c}{within-$Q$ variance share} & \\
\cmidrule(lr){4-5}\cmidrule(lr){6-7}
\textbf{Corpus / cell} & \textbf{Model family} & copy \% &
   DC & DC$_{\text{masked}}$ & DC & DC$_{\text{masked}}$ & RGV \\
\midrule
FRAMES        & GPT-OSS   & 82.1 & 0.68 & 0.49 & 0.35 & 0.52 & 0.46 \\
FRAMES        & Tongyi    & 85.2 & 0.59 & 0.37 & 0.49 & 0.57 & 0.57 \\
FRAMES        & GLM       & 87.8 & 0.78 & 0.75 & 0.16 & 0.30 & 0.31 \\
FRAMES        & Kimi      & 87.7 & 0.67 & 0.47 & 0.46 & 0.59 & 0.53 \\
FRAMES        & MiniMax   & 84.8 & 0.69 & 0.57 & 0.51 & 0.50 & 0.58 \\
BrowseComp-Plus & Tongyi  & 91.9 & 0.67 & 0.43 & 0.52 & 0.48 & 0.67 \\
BrowseComp-Plus & GPT-OSS & 91.9 & 0.73 & 0.43 & 0.18 & 0.42 & 0.59 \\
GAIA          & DeepSeek  & 89.7 & 0.70 & 0.72 & 0.16 & 0.24 & 0.48 \\
wiki-18 / HotpotQA & GPT-OSS & 78.4 & 0.82 & 0.49 & 0.13 & 0.29 & 0.49 \\
wiki-18 / FRAMES   & GPT-OSS & 81.9 & 0.79 & 0.44 & 0.12 & 0.28 & 0.41 \\
\bottomrule
\end{tabular}
\caption{\textbf{Masking copied tokens restores variance but not
validity.} \emph{copy \%} is the median fraction of generated tokens
that are copies. \emph{rollout AUC} separates judge-correct from
judge-wrong rollouts; \emph{within-$Q$ variance share} is the quantity
weighted voting consumes (\S\ref{sec:motivation}). Masking raises the
variance share in $8/10$ cells (up to $2.4\times$ on
BrowseComp-Plus\,/\,GPT-OSS, $0.18\!\to\!0.42$) while AUC \emph{falls}
in $9/10$.}
\label{tab:copymask}
\end{table*}

Three readings of Table~\ref{tab:copymask}. First, masking does
restore the spread, which confirms that copied tokens are what
flattens the score. Second, the restored spread carries no correctness
signal: AUC falls in nine of ten cells, and voting accuracy does not
systematically improve (up in $2$ cells, flat in $3$, down in $5$).
The discriminative information lives in precisely the tokens the
repair removes. Third, the retrieval-grounded score holds
$1.1$--$3.8\times$ (median $1.6\times$) more within-question variance
share than DeepConf in $10/10$ cells.

This is a negative result we consider more consequential than the
method itself: the contamination resists internal repair, so
\emph{any} consumer of agent token-confidence, whether for early stopping,
routing, abstention or reward shaping, inherits it, and no simple
adjustment recovers a usable weight.

\subsection{Intervention 2: removing the documents}

The second intervention removes the cause instead of the symptom. We
take $100$ BrowseComp-Plus\,/\,Tongyi-DeepResearch rollouts and
teacher-force the model over the \emph{same} answer tokens twice: once
with the retrieved documents present in context, and once with each
document replaced by the placeholder \texttt{[document removed]}
(trailing-context budget $24$K tokens). If the documents are what prop
up copied-token confidence, removing them should cost copied tokens
more than non-copy tokens.

\begin{center}\small
\setlength{\tabcolsep}{6pt}
\begin{tabular}{lcc}
\toprule
Token class & $\Delta$ logprob (mean) & $\Delta$ logprob (median) \\
\midrule
copied      & $-0.221$ & $-0.187$ \\
non-copy    & $-0.106$ & $-0.068$ \\
\bottomrule
\end{tabular}
\end{center}

\noindent
Copied tokens lose roughly twice as much log-probability, and the copy
drop exceeds the non-copy drop in $82$ of $100$ rollouts. Together
with the null controls of Figure~\ref{fig:why}(\subref{fig:why-b}) and
the token-subset robustness of Appendix~\ref{app:copy}, this closes the
causal chain: documents in context inflate the confidence of the
tokens copied from them.

% ======================================================================
\section{Prompt and answer-format sensitivity}
\label{app:prompt-format}

A grounding-based weight invites two worries: that a prompt could
inflate it, and that it could break on short answers. We test both
directly with gpt-oss-120b on BrowseComp-Plus, $150$ questions
$\times$ $8$ rollouts per arm, changing only one line of the prompt.

\begin{center}\footnotesize
\setlength{\tabcolsep}{3.5pt}
\begin{tabular}{lccccc}
\toprule
Arm & Single & SM & DC & RGV & \begin{tabular}{@{}c@{}}med.\\prose tok.\end{tabular} \\
\midrule
stock              & 47.3 & 53.3 & 54.7 & \textbf{56.0} & 254 \\
$+$ grounding instr.\ & 48.6 & 55.3 & 57.3 & \textbf{59.3} & 254 \\
$+$ entity-only    & 44.2 & 52.7 & 53.3 & \textbf{53.3} & 2 \\
\bottomrule
\end{tabular}
\end{center}

\paragraph{A prompt cannot buy weight.}
The grounding arm appends ``\emph{your final answer must be grounded
in the retrieved documents}''. The mean voted weight is unchanged
($0.091 \to 0.091$) and rollout-level AUC barely moves
($0.840 \to 0.844$). The reason is structural rather than empirical:
an instruction attached to the question reaches all $N$ rollouts of
that question equally, and weighted voting consumes only the
\emph{within-question ordering} of the weights, which a common shift
leaves intact. This is the same property that makes the weight robust
to a rollout being merely verbose (\S\ref{sec:method}).

\paragraph{Short answers degrade gracefully.}
The entity-only arm forces a bare entity as the final answer, taking
the median answer prose from $254$ tokens to $2$. At that length
prose-recall reduces to answer \emph{containment}: whether the
predicted entity occurs in the rollout's own documents. Rank AUC dips
from $0.71$ to $0.67$ on the truly short subset ($77\%$ of rollouts),
and $99\%$ of correct entities are anchored in their own retrieval
versus $80\%$ of wrong ones. Every method loses accuracy in this arm
because the model is weaker without its report format (single
$47.3 \to 44.2$); within it, the grounded vote still matches DeepConf
and leads simple majority.

This arm also isolates the design choice of \S\ref{sec:method}: a
symmetric, document-side-normalised overlap such as Jaccard collapses
to $42.0$ here, because a two-token answer can never cover a long
document. The answer-side denominator is what keeps the score
meaningful at this extreme.

% ======================================================================
\section{Selection versus voting}
\label{app:bon}

RGV is presented as a voting weight, but the same score can select a
single rollout. Separating the two tells us whether the gain comes
from the signal or from the aggregation frame. On the
BrowseComp-Plus\,/\,Tongyi-DeepResearch replay cell we compare
weighted voting against Best-of-$N$ selection (take the single
highest-scoring rollout), sub-sampling to $N \in \{1,2,4,8\}$ with
$60$ random partitions per $N$.

\begin{center}\footnotesize
\setlength{\tabcolsep}{2.6pt}
\begin{tabular}{cccccccc}
\toprule
$N$ & Single & SM & \begin{tabular}{@{}c@{}}vote\\RGV\end{tabular}
    & \begin{tabular}{@{}c@{}}vote\\DC\end{tabular}
    & \begin{tabular}{@{}c@{}}argmax\\RGV\end{tabular}
    & \begin{tabular}{@{}c@{}}argmax\\DC\end{tabular} & Oracle \\
\midrule
1 & 58.9 & 58.9 & 58.9 & 58.9 & 58.9 & 58.9 & 58.9 \\
2 & 58.9 & 58.9 & 63.3 & 60.3 & 63.3 & 60.2 & 70.1 \\
4 & 59.4 & 64.3 & 69.6 & 65.2 & 67.4 & 61.5 & 79.7 \\
8 & 59.2 & 67.3 & \textbf{74.0} & 70.0 & 69.3 & 60.7 & 86.7 \\
\bottomrule
\end{tabular}
\end{center}

\noindent
Two conclusions. RGV works as a standalone rollout-quality scorer:
argmax-RGV beats argmax-DeepConf at every $N$, and at $N{=}8$ it also
beats simple majority ($69.3$ vs $67.3$). But voting dominates
selection by $+4.7$ points at $N{=}8$, because argmax stakes the whole
answer on one rollout and is therefore fully exposed to the
well-grounded-but-wrong failure mode quantified in
\S\ref{sec:exp-error}; summing weights over an answer cluster averages
that risk away. Notably argmax-DeepConf \emph{decreases} from $N{=}4$
to $N{=}8$: with more rollouts to choose from, a flattened confidence
score is more likely to pick a confidently wrong one.

% ======================================================================
\section{Preliminary results on HLE}
\label{app:hle}

Our main evaluation focuses on search agent benchmarks where
retrieved documents are the primary evidence source
(\S\ref{sec:setup}). Here, we report one exploratory
cell on Humanity's Last Exam \citep{phan2025hle}, an
expert-reasoning benchmark where retrieval plays a minor role:

\begin{center}\small\setlength{\tabcolsep}{3pt}
\begin{tabular}{lccccc}
\toprule
& Single & SM & DC & \textbf{RGV} & Oracle \\
\midrule
OSS-120B & 25.6 & 31.2 & 31.8 & \textbf{32.5} & 49.0 \\
\bottomrule
\end{tabular}
\end{center}

\noindent RGV still outperforms DeepConf ($+0.7\%$), though the margin is smaller than on search-centric benchmarks. This is expected: HLE questions often require domain-expert reasoning beyond what retrieval supplies, so the grounding signal captures only part of the answer quality.

% ======================================================================
\section{Worked examples: prompts, thought spans, and clustering}
\label{app:examples}

This appendix shows the exact text the agent and the judge see,
plus illustrative excerpts of what a rollout produces. The
examples are lightly redacted (line breaks, bracketed
abbreviations) for readability; nothing functional is changed.

\subsection{Agent system prompt}

Below is the system prompt for BrowseComp-Plus (fixed-corpus
retrieval with \texttt{search} and \texttt{get\_doc}). For the
open-web benchmarks (BrowseComp, GAIA, FRAMES) the tool names are
\texttt{search} and \texttt{visit} and references to
\texttt{[docid]} become \texttt{[url]}; the rest of the prompt is
identical.

\begin{quote}\small\ttfamily\raggedright
You are a deep research assistant. Your core function is to
conduct thorough, multi-source investigations into any topic.
You must handle both broad, open-domain inquiries and queries
within specialized academic fields. For every request, synthesize
information from credible, diverse sources to deliver a
comprehensive, accurate, and objective response.\\[0.3em]
\# Tools\\
You have access to {\tt search} and {\tt get\_doc} tools. Use
{\tt search} to look up snippets; use {\tt get\_doc(docid)} to
read full documents you find via search. You may call tools many
times until you have gathered sufficient evidence.\\[0.3em]
\# Final answer format\\
When you have enough evidence, produce a comprehensive markdown
report containing:\\
-- A bolded direct answer to the question\\
-- {\tt \#\# Step-by-Step Reasoning and Evidence} section with
    sub-sections walking through each clue\\
-- A markdown table or bullet list summarizing how each
    criterion is satisfied\\
-- Inline citations with {\tt [docid]} for every nontrivial
    claim\\
Do not give a one-line answer; produce a structured research
report.
\end{quote}

\subsection{Force-final prompt}

Issued once when the token-count guard fires, max-iterations is
hit, the output budget is exhausted, or the API returns a
non-retryable error. Tools are disabled for this call.

\begin{quote}\small\ttfamily\raggedright
The conversation has been truncated. Based on the evidence you
have gathered so far, write the final markdown report now.
Use the format from the system prompt (bolded direct answer,
{\tt \#\# Step-by-Step Reasoning and Evidence}, a summary table,
inline {\tt [docid]} citations). End the report with a single
line of the form {\tt Confidence: NN\%}.
\end{quote}

\subsection{Verbalised-confidence solicitation}

Issued once after the rollout has produced a final answer (either
naturally or via force-final), at temperature $0$ with tools
disabled.

\begin{quote}\small\ttfamily\raggedright
Reply with ONLY a single integer between 0 and 100 representing
your confidence (in percent) that the answer you just gave is
correct. No words, no \%, no labels --- just the integer.
Example output: 73
\end{quote}

\subsection{Judge prompt}

The judge (Qwen3-32B at temperature $0$) receives the question,
the gold answer, and the rollout's final markdown report, and
decides correctness. We use the official BrowseComp-Plus
\texttt{GRADER\_TEMPLATE} for every dataset.

\begin{quote}\small\ttfamily\raggedright
Question: \{question\}\\
Gold answer: \{gold\_answer\}\\
Model response: \{model\_response\}\\[0.3em]
Read the model's response. First, extract the model's final
short answer in the {\tt extracted\_final\_answer} field
(or output {\tt None} if the model did not produce one). Then
decide whether the extracted answer matches the gold answer up
to obvious paraphrase / equivalent name. Output a JSON object
with fields {\tt extracted\_final\_answer}, {\tt reasoning},
and {\tt judge\_correct} $\in\{$ {\tt yes}, {\tt no} $\}$. Do
not output anything else.
\end{quote}

\subsection{Example thought span vs.\ answer prose}

A single rollout on the BrowseComp-Plus question
\textit{``There is an influencer whose career has lasted over a
decade\dots What is the full name of the influencer?''}
(gold answer: \emph{Tóke Makinwa}). The \emph{thought span} is
what DeepConf reads---chain-of-thought across all turns; the
\emph{answer prose} is what RGV reads---the assistant's output
on the final answer turn (no internal thought).

\paragraph{Thought span (excerpt, what DeepConf reads).}
\begin{quote}\small\ttfamily\raggedright
Need an influencer with a $>$10-year career who anchored a
2022 talk segment\dots The host is born in the 1980s and has
served as a host at major events. Possible candidates: Tope
Mark-Odigie, T\'oke Makinwa, Lilly Singh\dots T\'oke Makinwa
is listed as 2nd of 4 siblings on Wikipedia [docid 118],
became brand ambassador for Pulse Nigeria in 2016\dots Lilly
Singh is also 2nd of 4 but the 2022 talk-segment host
criterion fits T\'oke better\dots
\end{quote}

\paragraph{Answer prose (what RGV reads).}
\begin{quote}\small\ttfamily\raggedright
\textbf{T\'oke Makinwa.}\\[0.3em]
\#\# Step-by-Step Reasoning and Evidence\\
-- Long-career influencer with a YouTube vlog series:
  {\tt [docid 118]} records that T\'oke Makinwa launched the
  {\itshape Toke Moments} vlog in 2014, more than a decade
  after her 2004 radio debut.\\
-- 2022 talk-segment anchor: {\tt [docid 118]} notes she
  anchored the talk segment of {\itshape The Buzz}, hosted by
  Ebuka Obi-Uchendu (born 1982).\\
-- Second of four siblings: {\tt [docid 118]} lists her as the
  second of four.\\
-- 2016 ambassador for an MIT-founded brand: {\tt [docid 118]}
  records the Pulse Nigeria ambassadorship; Pulse's founder
  Tunde Kehinde is an MIT Sloan alumnus.
\end{quote}

\noindent
$\mathcal{T}(P_i)$ is built from the \emph{Answer prose} block
only; DeepConf's per-token confidences are computed over the
\emph{Thought span} only. The two operate on disjoint parts of
the rollout's output.

\subsection{Example clustering on one question}

For the same question, the eight rollouts produced the
following predicted-answer strings, normalised under the strict
rule:

\begin{center}\footnotesize
\setlength{\tabcolsep}{3pt}
\begin{tabular}{r l l c}
\toprule
\# & raw predicted answer & normalised & cluster \\
\midrule
1 & T\'oke Makinwa.             & toke makinwa   & A \\
2 & The answer is Toke Makinwa. & toke makinwa   & A \\
3 & Lilly Singh                 & lilly singh    & B \\
4 & Lilly Singh.                & lilly singh    & B \\
5 & Lilly Singh                 & lilly singh    & B \\
6 & Toke Makinwa                & toke makinwa   & A \\
7 & Shannon LaNier              & shannon lanier & C \\
8 & Bhuvan Bam                  & bhuvan bam     & D \\
\bottomrule
\end{tabular}
\end{center}

\noindent
Strict clustering recovers four clusters with $|A|{=}3$,
$|B|{=}3$, $|C|{=}|D|{=}1$. Simple-majority ties A and B at
weight $3$; under DeepConf the eight rollouts all receive
nearly identical weights and the tie persists; under RGV the
three A-cluster rollouts carry the largest prose-recall (their
answer prose lexically aligns with the \textit{Tóke Makinwa}
Wikipedia entry the agent actually fetched), so the weighted
vote selects cluster A.

% ======================================================================
\section{Success and failure cases}
\label{app:error-cases}

We first show three representative \emph{successes}---minority-correct
questions where RGV rescues the answer that DeepConf misses---then
five \emph{failures} exhibiting the ``well-grounded but wrong''
pattern of \S\ref{sec:exp-error}.

\subsection{Success cases: RGV rescues the correct minority}

DeepConf gives all rollouts near-identical confidence and follows
the wrong majority; RGV elevates the correctly grounded rollout.

\paragraph{Success 1 (qid 301): royal family identification.}
\emph{Q:} ``Name the royal family a certain individual's spouse was
born into\ldots'' (multiple biographical constraints).\\
\emph{Gold:} Nwoko.

\begin{center}\small\setlength{\tabcolsep}{3pt}
\begin{tabular}{clccl}
\toprule
 & & RGV & DC & Predicted answer \\
\midrule
r2 & \ding{51} & \textbf{0.184} & 6.54 & Nwoko Royal Family \\
r6 & \ding{51} & 0.118 & 6.66 & Nwoko Royal Family \\
r7 & \ding{55} & 0.071 & 6.46 & Greek royal family \\
r1 & \ding{55} & 0.026 & 6.70 & Greek royal family \\
r5 & \ding{55} & 0.024 & 6.68 & British royal family \\
r0 & \ding{55} & 0.009 & 6.75 & House of Orange-Nassau \\
r4 & \ding{55} & 0.008 & 6.59 & House of Windsor \\
\bottomrule
\end{tabular}
\end{center}
\noindent Seven distinct answers. The two correct rollouts retrieve
the Nwoko family page and echo it in their prose
($\text{RGV}\geq0.118$); all wrong rollouts name famous European
royal families without retrieving relevant evidence (RGV~$\leq0.071$).
DeepConf range: $6.46$--$6.75$, unable to discriminate.

\paragraph{Success 2 (qid 424): architect identification.}
\emph{Q:} ``Name the architect who was a WWII veteran and TV
broadcaster, designed a building completed 1977--1987\ldots''\\
\emph{Gold:} Raffaele Contigiani.

\begin{center}\small\setlength{\tabcolsep}{3pt}
\begin{tabular}{clccl}
\toprule
 & & RGV & DC & Predicted answer \\
\midrule
r4 & \ding{51} & \textbf{0.152} & 6.98 & Raffaele Contigiani \\
r6 & \ding{51} & 0.133 & 6.72 & Raffaele Contigiani \\
r1 & \ding{55} & 0.095 & 6.75 & Victor Alfred Lundy \\
r3 & \ding{55} & 0.092 & 6.99 & Denys Lasdun \\
r2 & \ding{55} & 0.073 & 6.76 & Denys Lasdun \\
r0 & \ding{55} & 0.018 & 6.48 & Denys Lasdun \\
\bottomrule
\end{tabular}
\end{center}
\noindent Three wrong rollouts converge on Denys Lasdun (a famous
British architect who fits some but not all constraints). The
correct rollouts retrieve Contigiani's Italian biography and echo
it in prose, producing a clear RGV separation ($0.152$ vs.\ $0.095$).

\paragraph{Success 3 (qid 56): movie identification.}
\emph{Q:} ``Movie with a ReFrame Stamp, 2018--2023, about a
festival, with a cast member who appeared in a heist film\ldots''\\
\emph{Gold:} Last Christmas.

\begin{center}\small\setlength{\tabcolsep}{3pt}
\begin{tabular}{clccl}
\toprule
 & & RGV & DC & Predicted answer \\
\midrule
r2 & \ding{51} & \textbf{0.115} & 7.53 & Last Christmas \\
r5 & \ding{55} & 0.028 & 7.96 & Crazy Rich Asians \\
r4 & \ding{55} & 0.026 & 7.54 & Crazy Rich Asians \\
r3 & \ding{55} & 0.026 & 7.69 & Crazy Rich Asians \\
r7 & \ding{55} & 0.018 & 7.78 & Rifkin's Festival \\
r1 & \ding{55} & 0.018 & 7.37 & Rifkin's Festival \\
r6 & \ding{55} & 0.016 & 7.70 & A Simple Favor \\
\bottomrule
\end{tabular}
\end{center}
\noindent Three rollouts vote for Crazy Rich Asians with high
DeepConf ($7.54$--$7.96$); DeepConf selects this wrong majority.
RGV: the single correct rollout retrieves the film's page ($0.115$
vs.\ ${\leq}0.028$) and wins.

\subsection{Failure cases: well-grounded but wrong}

\paragraph{Failure 1 (qid 1185): cricket match.}
\emph{Q:} ``Match number, tournament, year satisfying nine batting
constraints\ldots''\\
\emph{Gold:} 31st match, IPL 2013.

\begin{center}\small\setlength{\tabcolsep}{3pt}
\begin{tabular}{clccl}
\toprule
 & & RGV & DC & Predicted answer \\
\midrule
r7 & \ding{55} & \textbf{0.235} & 6.93 & 2019 ICC World Cup \\
r5 & \ding{55} & 0.174 & 6.98 & Bangladesh vs South Africa \\
r2 & \ding{51} & 0.141 & 7.30 & 31st match, IPL 2013 \\
\bottomrule
\end{tabular}
\end{center}
\noindent r7 retrieves extensive World Cup documentation and
echoes cricket statistics ($\text{RGV}=0.235$), outscoring the
correct rollouts ($0.165$, $0.141$) despite naming the wrong
tournament.

\paragraph{Failure 2 (qid 790): writer's work identification.}
\emph{Q:} ``A writer born in July, article posted April 2023,
smoked on Christmas\ldots name the work.''\\
\emph{Gold:} ``Manos''.

\begin{center}\small\setlength{\tabcolsep}{3pt}
\begin{tabular}{clccl}
\toprule
 & & RGV & DC & Predicted answer \\
\midrule
r0 & \ding{55} & \textbf{0.229} & 6.96 & ``Killing Joke'' \\
r3 & \ding{51} & 0.206 & 6.68 & ``Manos'' \\
r1 & \ding{55} & 0.094 & 6.71 & (different work) \\
r7 & \ding{55} & 0.008 & 6.70 & ``mouse'' \\
r4 & \ding{55} & 0.000 & 6.80 & ``girls'' \\
r5 & \ding{55} & 0.000 & 6.83 & ``Thugs'' \\
\bottomrule
\end{tabular}
\end{center}
\noindent r0 retrieves an article about the same writer but
identifies a different work ($0.229$ vs.\ $0.206$). Both rollouts
ground their answer in the writer's bibliography; the wrong one
happens to echo a more frequently cited title.

\paragraph{Failure 3 (qid 773): clothing colour.}
\emph{Q:} ``A child reported missing multiple times 2014--2018.
What colour shirt in the 2018 police description?''\\
\emph{Gold:} Red.

\begin{center}\small\setlength{\tabcolsep}{3pt}
\begin{tabular}{clccl}
\toprule
 & & RGV & DC & Predicted answer \\
\midrule
r7 & \ding{55} & \textbf{0.209} & 7.41 & White t-shirt \\
r3 & \ding{55} & 0.190 & 7.27 & (different colour) \\
r4 & \ding{55} & 0.180 & 7.68 & (different colour) \\
r2 & \ding{55} & 0.179 & 7.46 & (different colour) \\
r5 & \ding{55} & 0.166 & 7.88 & (different colour) \\
r6 & \ding{55} & 0.142 & 8.05 & (different colour) \\
r1 & \ding{51} & 0.109 & 7.11 & Red \\
r0 & \ding{55} & 0.000 & 6.45 & (different colour) \\
\bottomrule
\end{tabular}
\end{center}
\noindent All seven wrong rollouts retrieve articles about the same
missing-person case and echo detailed police descriptions, receiving
higher RGV scores ($0.142$--$0.209$) than the single correct
rollout ($0.109$). The correct answer (``Red'') is a single word
with minimal lexical overlap; wrong rollouts produce longer prose
quoting extensively from the retrieved reports.

\paragraph{Pattern and future directions.}
Each failure shares a structure: the wrong rollout retrieves
documents about the right \emph{domain} (same writer, same sport,
same missing-person case) but identifies the wrong \emph{specific
entity} within that domain. RGV measures whether the answer is
anchored in retrieved evidence; it cannot verify that the evidence
answers the specific question asked. Closing this gap likely
requires moving beyond surface-level lexical overlap toward
\emph{semantic} grounding checks, for example, entailment
verification between the question constraints and the retrieved
passages, or cross-referencing the answer against multiple
independent sources. Such extensions could complement RGV's
efficiency with deeper reasoning at a modest additional cost.

% ======================================================================
\section{Artifacts, licences, and intended use}
\label{app:licences}

Table~\ref{tab:licences} lists every external artifact this work uses,
with the licence under which it is released. We used each within the
research use its original release specifies.

\begin{table*}[t]
\centering\footnotesize
\setlength{\tabcolsep}{4pt}
\renewcommand{\arraystretch}{1.1}
\begin{tabular}{@{}p{5.0cm} l l p{5.4cm}@{}}
\toprule
\textbf{Artifact} & \textbf{Type} & \textbf{Licence} & \textbf{Note on use} \\
\midrule
BrowseComp-Plus \citep{chen2025browsecompplus} & benchmark & MIT & questions, corpus, gold-evidence qrels \\
BrowseComp \citep{wei2025browsecomp}           & benchmark & MIT & via \texttt{openai/simple-evals} \\
FRAMES \citep{krishna2024frames}               & benchmark & Apache-2.0 & questions and gold answers \\
GAIA \citep{mialon2023gaia}                    & benchmark & gated, none stated & validation split; redistribution not permitted \\
HLE \citep{phan2025hle}                        & benchmark & MIT & authors ask that it not be re-uploaded \\
HotpotQA \citep{yang-etal-2018-hotpotqa}       & benchmark & CC BY-SA 4.0 & Appendix~\ref{app:weak-retrieval} only \\
wiki-18 \citep{jin2025searchr1} & corpus & CC BY-SA & Wikipedia derivative; Appendix~\ref{app:weak-retrieval} only \\
\midrule
gpt-oss-120b \citep{openai2025gptoss}          & model & Apache-2.0 & open weights, served via API \\
Tongyi-DeepResearch-30B-A3B\newline \citep{tongyideepresearch2025} & model & Apache-2.0 & open weights, served locally (vLLM, FP8) \\
MiniMax-M2.7 \citep{minimax2025m1}             & model & per model card & served via API \\
GLM-5.1 \citep{zai2026glm5}                    & model & per model card & served via API \\
Kimi-K2.5 \citep{moonshot2025kimik2}           & model & per model card & served via API \\
Qwen3-32B                                      & judge & Apache-2.0 & served locally (vLLM) \\
\texttt{e5-base-v2} \citep{wang2022e5}         & encoder & MIT & Appendix~\ref{app:lex-metrics} baseline \\
\bottomrule
\end{tabular}
\caption{External artifacts and their licences.}
\label{tab:licences}
\end{table*}

\paragraph{What we release.}
We release our code, the full rollout trajectories, the per-token
log-probabilities, and the judge outputs for every cell reported in
this paper. Our own contributions in that release, namely the trajectories,
scores, judge labels and derived annotations, are made available
under CC BY 4.0. Text that the environment returned (retrieved corpus
passages and fetched web pages) remains under its original terms and
is redistributed with its source identifiers retained; we provide a
contact for removal requests.

\paragraph{Benchmarks we do not redistribute.}
Two of the benchmarks we evaluate on restrict redistribution, and we
respect those terms. GAIA is access-gated and states that its
validation split may not be reshared; HLE's authors ask that the
benchmark not be publicly re-uploaded, to protect it from
contamination. For these two, our release carries question
identifiers rather than question and gold-answer text, so that
researchers who have obtained the benchmarks through their official
channels can join our records against them, while the release itself
does not republish the benchmark.

\paragraph{Data characteristics.}
All questions and answers are in English. The benchmarks consist of
factual information-seeking questions and contain no personal or
offensive content by construction. The agents' retrieved documents
come from a fixed corpus (BrowseComp-Plus, wiki-18) or the live web
via a commercial search API (BrowseComp, GAIA, FRAMES); we scan the
released tool outputs for credentials and personally identifying
information before publication.

\end{document}